 \documentclass[tablecaption=bottom,pmlr,nosubfloats]{jmlr}

\usepackage{bbm}
\usepackage{algorithm}
\usepackage{algorithmic}
\usepackage{tikz}
\usepackage{multirow}
\usepackage{arydshln}
\usepackage{float}
\DeclareMathOperator*{\argmax}{arg\,max}
\usepackage{subcaption}

\jmlrvolume{X}
\jmlryear{2021}
\jmlrsubmitted{XXXX}
\jmlrpublished{XXXX}
\jmlrworkshop{Pre-print} % W&CP title
\jmlrproceedings{XXXXXX}{XXXXXX}

\title[Domain Conditional Predictors for Domain Adaptation]{Domain Conditional Predictors for Domain Adaptation}

  \author{\Name{Jo\~ao Monteiro\nametag{\thanks{Work performed while Jo\~ao Monteiro was interning at Google.}}} \Email{joao.monteiro@inrs.ca}\\
  \addr INRS-EMT, Universit\'e du Qu\'ebec
  \AND
  \Name{Xavier Gibert} \Email{xgibert@google.com}\\
  \Name{Jianqiao Feng} \Email{jianqiaofeng@google.com}\\
  \Name{Vincent Dumoulin} \Email{vdumoulin@google.com}\\
  \Name{Dar-Shyang Lee} \Email{dsl@google.com}\\
  \addr Google
  }

\begin{document}

\maketitle

\begin{abstract}

Learning guarantees often rely on assumptions of i.i.d. data, which will likely be violated in practice once predictors are deployed to perform \emph{real-world} tasks. Domain adaptation approaches thus appeared as a useful framework yielding extra flexibility in that distinct train and test data distributions are supported, provided that other assumptions are satisfied such as \emph{covariate shift}, which expects the conditional distributions over labels to be independent of the underlying data distribution. Several approaches were introduced in order to induce generalization across varying train and test data sources, and those often rely on the general idea of domain-invariance, in such a way that the data-generating distributions are to be disregarded by the prediction model. In this contribution, we tackle the problem of generalizing across data sources by approaching it from the opposite direction: we consider a conditional modeling approach in which predictions, in addition to being dependent on the input data, use information relative to the underlying data-generating distribution. For instance, the model has an explicit mechanism to adapt to changing environments and/or new data sources. We argue that such an approach is more generally applicable than current domain adaptation methods since it does not require extra assumptions such as covariate shift and further yields simpler training algorithms that avoid a common source of training instabilities caused by minimax formulations, often employed in domain-invariant methods.

\end{abstract}


\section{Introduction}

Common generalization guarantees used to motivate supervised learning approaches under the empirical risk minimization framework (ERM) rely on the assumption that data is collected independently from a fixed underlying distribution. Such an assumption, however, is not without shortcomings; for instance: (\textbf{i})-i.i.d. requirements are \emph{unverifiable} \citep{langford2005tutorial} in the sense that, given a data sample and no access to the distribution it was observed from, one cannot tell whether such a sample was collected independently, and (\textbf{ii})-the i.i.d. assumption is \emph{unpractical} since in several scenarios the conditions under which data is collected will likely change relative to when training samples were observed and, as such, generalization cannot be expected. Yet another practical limitation given by the lack of robustness against distribution shifts in common predictors is the fact that one cannot benefit from data sources that differ from those against which such predictors will be tested. In some situations, for example, data can be collected from inexpensive simulations, but generalization to real data depends on how similar the synthetic data distribution is to the data distribution of interest.

Several approaches have been consequently introduced with the goal of relaxing requirements of i.i.d. data to some extent. For instance, \emph{domain adaptation approaches} \citep{ben2007analysis} assume the existence of two distributions: the source distribution -- which contains the bulk of the training data -- and the target distribution -- which corresponds to the test-time data distribution. While the domain adaptation setting enlarged the scope of the standard empirical risk minimization framework by enabling the use of predictors even when a distribution other than the one used for training is considered, a particular \emph{target} is expected to be defined at training time, often with unlabeled examples, and nothing can be guaranteed for distributions other than that particular \emph{target}, which renders such setting still unpractical since unseen variations in the data are possible during test. More general settings were introduced considering a larger set of supported target distributions while not requiring access to any target data during training \citep{albuquerque2019generalizing}. However, such approaches, including domain adaptation techniques discussed so far, despite of relaxing the i.i.d. requirement, still require other assumptions to be met such as \emph{covariate shift} (c.f. sec. \ref{sec:related_work} for a definition).

As will be further discussed in Section~\ref{sec:related_work}, a common feature across a number of approaches enabling some kind of out-of-distribution generalization is that they rely on some notion of invariance, be it at the feature level \citep{ganin2016domain,albuquerque2019generalizing}, in the sense that the domains' data distributions cannot be discriminated after being mapped to some intermediate features by a feature extractor, or in the predictor level \citep{arjovsky2019invariant,krueger2020out} in which case one expects distribution shifts will have little effect over predictions. In this contribution, the research question we pose to ourselves is as follows: \emph{can one leverage contextual information to induce generalization to novel data sources?} We thus take an alternative approach relative to previous work and propose a framework where the opposite direction is considered in order to tackle the limitations discussed above, i.e. instead of filtering out the domain influence over predictions, we explore approaches where such information is leveraged as a source of context on which predictions can be conditioned. We refer to predictors resulting of such an approach as \emph{domain conditional predictors}. We argue such a method includes the following advantages compared to settings seeking invariance:

\begin{enumerate}
    \item Training strategies yielding domain conditional predictors do not rely on minimax formulations, often employed in domain invariant approaches where a domain discriminator is trained against a feature extractor. Such formulations are often observed to be source of training instabilities which do not appear in the setting considered herein.
    \item The proposed setting does not rely on the covariate shift assumption since it considers multiple inputs, i.e. for a fixed input data instance, any prediction can be obtained through variations of the conditioning variable.
    \item The proposed setting has a larger scope when compared to domain-invariant approaches in the sense that it can be used to perform inferences regarding the domains it observed during training.
\end{enumerate}

The remainder of this paper is organized as follows: related literature is discussed in Section~\ref{sec:related_work} along with background information and notation, while the proposed approach is presented in Section~\ref{sec:method}. The planned experimental setup is discussed in Section~\ref{sec:evaluation} and proof-of-concept results are reported in Section~\ref{sec:experiments}. The complete evaluation is reported in Section~\ref{sec:complete_evaluation}, while conclusions as well as future directions are drawn in Section~\ref{sec:conclusion}.

\section{Background and Related Work}
\label{sec:related_work}

\subsection{Domain adaptation guarantees and domain invariant approaches}

Assume $(x, y)$ represents instances from $\mathcal{X} \times \mathcal{Y}$, where $\mathcal{X} \subseteq \mathbb{R}^D$ is the data space while $\mathcal{Y}$ is the space of labels, and $\mathcal{Y}$ will be a discrete set in the cases we consider. Furthermore, consider a deterministic labeling function, denoted $f:\mathcal{X} \mapsto \mathcal{Y}$, is such that $y=f(x)$. We will refer to domains as the pairs given by a marginal distribution over $\mathcal{X}$, denoted by $\mathcal{D}$, and a labeling function $f$. We further consider a family of candidate predictors $\mathcal{H}$ where $h \in \mathcal{H} : \mathcal{X} \mapsto \mathcal{Y}$. For a particular predictor $h$, we use the standard definition of risk $R$, which is its expected loss:

\begin{equation}
    R_{\mathcal{D}}[h] = \mathbb{E}_{x \sim \mathcal{D}} \ell [h(x), f(x)],
\end{equation}
where the loss $\ell:\mathcal{Y} \times \mathcal{Y} \rightarrow \mathbb{R}_{+}$ indicates how different $h(x)$ and $f(x)$ are (e.g. the 0-1 loss for $\mathcal{Y}=\{0,1\}$).

\cite{ben2007analysis} showed that the following bound holds for the risk on a target domain $\mathcal{D}_T$ depending on the risk measured on the source domain $\mathcal{D}_S$:

\begin{equation}\label{eq:bound_da}
    R_{\mathcal{D}_T}[h] \leq R_{\mathcal{D}_S}[h] + d_{\mathcal{H}}[\mathcal{D}_S, \mathcal{D}_T] + \lambda,
\end{equation}
and the following details are worth highlighting regarding such result: (\textbf{i})-the term $\lambda$, as discussed by \cite{zhao2019learning}, accounts for differences between the labeling functions in the source and target domains, i.e. in the more general case the label $y'$ of a particular data point $x'$ depends on the underlying domain it was observed from. The \emph{covariate shift} assumption thus considers the more restrictive case where labeling functions match across domains, zeroing out $\lambda$ and tightening the bound shown above. (\textbf{ii})-the term $d_{\mathcal{H}}[\mathcal{D}_S,\mathcal{D}_T]$ corresponds to the discrepancy measure in terms of the $\mathcal{H}$-divergence (c.f. definition in \citep{ben2007analysis}) as measured across the two considered domains.

The covariate shift assumption thus induces a setting where generalization can be expected if the considered domains lie close to each other in the $\mathcal{H}$-divergence sense. Such setting motivated the domain invariant approaches appearing across a number of recent domain adaptation methods \citep{ganin2016domain,albuquerque2019generalizing,bhattacharya2019generative,bashivan2020adversarial} where a feature extractor is forced to ignore domain-specific cues from the data and induce a low discrepancy across domains, enabling generalization. A similar direction was recently proposed to define invariant predictors instead of invariant representations in \citep{arjovsky2019invariant,krueger2020out,ahuja2020invariant}. In such setting, while data representations might still be domain-dependent, one seeks predictors that disregard domain factors in the sense that their predictions are not affected by changes in the underlying domains.

\subsection{Conditional modeling}

The problem of conditional modeling, i.e. that of defining models of a conditional distribution where some kind of contextual information is considered, appears across several areas. In this work, we choose to use a conditioning approach commonly referred to as FiLM, which belongs to a family of feature-wise conditioning methods that is widely used in the literature \citep{dumoulin2018feature-wise}. FiLM layers were introduced by \cite{perez2017film} and employed to tackle multi-modal visual reasoning tasks. Another setting where FiLM layers have been shown effective is few-shot learning. This is the case, for instance, of TADAM \citep{oreshkin2018tadam}, CNAPs \citep{requeima2019fast}, and CAVIA \citep{zintgraf2019fast} where FiLM layers are used to enable adapting a global cross-task model to particular tasks. Moreover, the few-shot classification setting under domain shift is tackled in \citep{Tseng2020Cross-Domain}, where feature-wise transformations are used as a means to diversify data and artificially create new domains at training time.

In further detail, FiLM layers consist of a per-feature affine transformation where its parameters are themselves a function of the data, as defined in the following:

\begin{equation}
\label{eq:film_operator}
\text{FiLM}(x,z) = \gamma(z)F(x)+\beta(z),    
\end{equation}
where $F(x)$ represents features extracted from an input denoted $x$, while $\gamma(z)$ and $\beta(z)$ are arbitrary functions of some available conditioning information from the data and represented by $z$ (e.g. $x$ corresponded to images and $z$ represented text in the original paper). $\gamma$ and $\beta$ were thus parameterized by neural networks and trained jointly with the main model, function of $x$.

An approach similar to FiLM was employed by \cite{karras2019style} for the case of generative modeling. A conditioning layer consisting of adaptive instance normalization layers \citep{huang2017arbitrary} was used to perform conditional generation of images providing control over the style of the generated data. Moreover, in \citep{prol2018cross} FiLM layers were applied in order to condition a predictor on entire classification tasks. The goal in that case was to enable adaptable predictors for few-shot classification of novel classes.

Other applications of such a framework include, for instance, conditional language modeling. In \citep{keskar2019ctrl} for example, deterministic codes are given at train time indicating the style of a particular corpus. At test time, one can control the sampling process by giving each such code as an additional input and generate outputs corresponding to a particular style. In the case of applications to speech recognition, a common approach for acoustic modelling is to augment acoustic features with speaker-dependent representations so that the model can account for factors that are specific to the underlying speaker such as accent and speaking speed \citep{peddinti2015time}. Representations such as i-vectors \citep{dehak2010front} are then combined with spectral features at every frame, and the combined representations are projected onto a mixed subspace, learned prior to training the acoustic model.

\section{Domain conditional predictors}
\label{sec:method}

The setting we consider consists in designing models that parameterize a conditional categorical distribution which simultaneously relies on data as well as on domain information. We then assume training data comes from multiple source domains. Each example has a label that indicates which domain it belongs to. Additional notation is introduced to account for that, in which case we denote the set of domain labels by $\mathcal{Y}_{\mathcal{D}}$. We then consider two models given by:

\begin{enumerate}
    \item $M_{domain} : \mathcal{X} \mapsto \Delta^{|\mathcal{Y}_{\mathcal{D}}|-1}$ maps a data instance $x$ onto the $|\mathcal{Y}_{\mathcal{D}}|-1$ probability simplex that defines the following categorical conditional distribution: $P(\mathcal{Y}_{\mathcal{D}}|x)=M_{domain}(x)$.
    
    \item $M_{task} : \mathcal{X} \times \mathbb{R}^d \mapsto \Delta^{|\mathcal{Y}|-1}$, where an extra input represented by $z \in \mathbb{R}^d$ is a conditioning variable expected to carry domain information. $M_{task}$ maps a data instance $x$ and its corresponding $z$ onto the $|\mathcal{Y}|-1$ probability simplex, thus defining the following categorical conditional distribution: $P(\mathcal{Y}|x,z)=M_{task}(x,z)$.
\end{enumerate}

We implement both $M_{domain}$ and $M_{task}$ using neural networks, and training is carried out with simultaneous maximum likelihood estimation over $P(\mathcal{Y}_{\mathcal{D}}|x)$ and $P(\mathcal{Y}|x,z)$ so that the training objective $\mathcal{L}=(1-\lambda)\mathcal{L}_{task}+\lambda\mathcal{L}_{domain}$ is defined by the sum of the multi-class cross-entropy losses defined over the set of task and domains labels, respectively, and $\lambda \in [0,1]$ is a hyperparameter that controls the importance of each loss term during training. Moreover, $z$ is given by the output of some inner layer of $M_{domain}$, since $z$ is expected to contain domain-dependent information. A training procedure is depicted in Algorithm \ref{alg:training}.

In order for $M_{task}$ to be able to use the domain conditioning information made available through $z$, we make use of FiLM layers represented by:

\begin{equation}
    FiLM^k(x^{k-1}, z)=(W^k_1z+b^k_1)x^{k-1}+(W^k_2z+b^k_2),
\end{equation}
where $k$ indicates a particular layer within $M_{task}$, and $x^{k-1}$ corresponds to the output of the previous layer. $W^k_1$, $b^k_1$, $W^k_2$, and $b^k_2$ correspond to the conditioning parameters trained along with the complete model.

\vspace{1cm}
\begin{algorithm}[]
\caption{Training procedure.}
\label{alg:training}
\begin{algorithmic}
   %\REQUIRES{a}
   \STATE $M_{task}, M_{domain} = InitializeModels()$
   \REPEAT
   \STATE $x, y, y_{\mathcal{D}} = SampleMinibatch()$
   \STATE $y'_{\mathcal{D}}, z = M_{domain}(x)$
   \STATE $y' = M_{task}(x,z)$
   \STATE $\mathcal{L}=\mathcal{L}_{task}(y',y)+\mathcal{L}_{domain}(y'_{\mathcal{D}},y_{\mathcal{D}})$
   \STATE $M_{task}, M_{domain} = UpdateRule(M_{task}, M_{domain}, \mathcal{L})$
   \UNTIL{Maximum number of iterations reached}
   \STATE \textbf{return} $M_{task}, M_{domain}$
\end{algorithmic}
\end{algorithm}
\vspace{1cm}

At test time, two distinct classifiers can be defined such as the task predictor given by:

\begin{equation}
    \argmax_{i \in [|\mathcal{Y}|]} M_{task}(x,z)_i,
\end{equation}
or the domain predictor defined by:
\begin{equation}
    \argmax_{j \in [|\mathcal{Y}_{\mathcal{D}}|]} M_{domain}(x)_j,
\end{equation}
thus enabling extra prediction mechanisms compared to methods that remove domain information.

\section{Planned evaluation and results}
\label{sec:evaluation}

In this section, we list the datasets we consider for the evaluation along with baseline methods, ablations, and the considered variations of conditioning approaches.

\subsection{Datasets}

Evaluations are to be performed on a subset of the following well-known domain generalization benchmarks:

\begin{itemize}
\item PACS \citep{li2017deeper}: It consists of 224x224 RGB images distributed into 7 classes and originated from four different domains: Photo (P), Art painting (A), Cartoon (C), and Sketch (S).

\item VLCS \citep{fang2013unbiased}: VLCS is composed by 5 overlapping classes of objects obtained from the following datasets: VOC2007 \citep{everingham2010pascal}, LabelMe \citep{russell2008labelme}, Caltech-101 \citep{griffin2007caltech}, and SUN \citep{choi2010exploiting}.

\item OfficeHome \citep{venkateswara2017Deep}: This dataset consists of images from the following domains: artistic images, clip art, product images and natural images. Each domain contains images of 65 classes.

\item DomainNet \citep{peng2019moment}: DomainNet contains examples of 224x224 RGB images corresponding to 345 classes of objects across 6 distinct domains.
\end{itemize}

\paragraph{Evaluation metric} Across all mentioned datasets, we follow the \emph{leave-one-domain-out} evaluation protocol such that data from $|\mathcal{Y}_{\mathcal{D}}|-1$ out of the $|\mathcal{Y}_{\mathcal{D}}|$ available domains are used for training, while evaluation is carried out on the data from the left out domain. This procedure is repeated for all available domains, and once each domain is left out, the \emph{average  top-1 accuracy} is the evaluation metric under consideration. Moreover, in order to provide  comparisons with significance, performance is to be reported in terms of confidence intervals obtained from independent models trained with different random seeds.

\subsection{Baselines}

The main aspect under investigation within this work is whether one can leverage domain information rather than removing or disregarding it such as in typical settings. Our main baselines then correspond to two settings where some kind of domain invariance is enforced: \emph{domain-adversarial approaches} and \emph{invariant predictors}. We specifically consider DANN \citep{ganin2016domain} and G2DM \citep{albuquerque2019generalizing} corresponding to the former, while IRM \citep{arjovsky2019invariant} and Rex \citep{krueger2020out} are considered for the latter. Additionally, two further baselines are evaluated: an \emph{unconditional model} in the form of a standard classifier that disregards domain labels as well as a model replacing $M_{domain}$ by a standard embedding layer\footnote{For this case, evaluation is performed in-domain, i.e. fresh data from the same domains observed at training time are used for testing.}.

\subsection{Ablations}

In order to investigate different sources of potential improvement, we will drop the domain classification term of the loss ($\mathcal{L}_{domain}$), obtaining a predictor with the same capacity as the proposed model while having no explicit mechanism for conditional modeling. A drop in performance should serve as evidence that the conditioning approach yields improvement. Moreover, similarly to ablations performed in the original FiLM paper \citep{perez2017film}, we plan on evaluating cases where scaling and offset parameters (i.e. $\gamma$ and $\beta$ as indicated in Eq. \ref{eq:film_operator}) are all set to 1 or 0, indicating which parameter set is more important for the conditioning strategy to be effective.

\subsection{Further evaluation details}

As discussed in previous work \citep{albuquerque2019generalizing, krueger2020out, gulrajani2020search}, the chosen validation data source used to implement model selection and stopping criteria significantly affects the performance of domain generalization approaches. We remark that, in this work, the access model to left out domains is such that no access to target data is allowed for model selection or hyperparameter tuning. We thus only use in-domain validation data. However, we further consider the so-called ``privileged'' variants of both our models and baselines in the sense that they are given access to target data. In doing so, we can get a sense of the gap in performance observed across the settings.

\section{Proof-of-concept evaluation}
\label{sec:experiments}

\begin{figure}[h]
\centering
\begin{subfigure}[b]{0.24\textwidth}
\centering
\includegraphics[width=3cm]{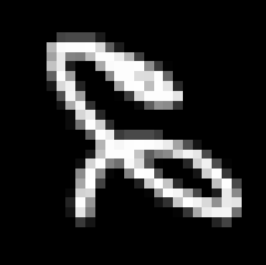}
\caption{Horiz. flip.}
\label{fig:hflip}
\end{subfigure}
\begin{subfigure}[b]{0.24\textwidth}
\centering
\includegraphics[width=3cm]{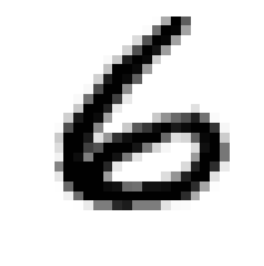}
\caption{Flip colors.}
\label{fig:colorflip}
\end{subfigure}
\begin{subfigure}[b]{0.24\textwidth}
\centering
\includegraphics[width=3cm]{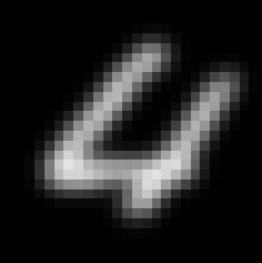}
\caption{Blurring.}
\label{fig:blur}
\end{subfigure}
\begin{subfigure}[b]{0.24\textwidth}
\centering
\includegraphics[width=3cm]{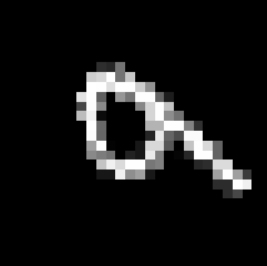}
\caption{Rotation.}
\label{fig:rotation}
\end{subfigure}
\caption{Examples of transformations used to simulate different data sources out of MNIST.}
\end{figure}

We used MNIST to perform validation experiments and considered different domains simulated through transformations applied on training data. Considered such transformations are as follows: (\textbf{i})-horizontal flip of digits, (\textbf{ii})-switching colors between digits and background, (\textbf{iii})-blurring, and (\textbf{iv})-rotation. Examples from each transformation are shown in Figures \ref{fig:hflip}-\ref{fig:rotation}. Test data corresponds to the standard test examples without any transformation. In-domain performance is further assessed by applying the same transformations on the test data.
Two baselines are considered in this set of experiments consisting of an unconditional model as well as a domain adversarial approach similar to DANN. For the case of the adversarial baselines, training is carried out so that alternate updates are performed to jointly train a task classifier and a domain classifier. The task classifier trains to minimize its classification loss and further maximizes the entropy of the domain classifier aiming to enforce domain invariance in the representations extracted after its convolutional layer. The domain classifier trains to correctly classify domains. Two ablations are also considered. The first one consists of a conditional model with learned domain-level context variables used for conditioning, in which case the conditioning model is replaced by an embedding layer. Such model can only be evaluated in in-domain test data. Additionally, we consider the ablation described above so that the domain classification term of the loss is dropped. For both the case of baselines and conditional approaches, classifiers are implemented as two-layered ReLU activated convolutional networks. Moreover, in the case of conditional models, $M_{domain}$ is given by a single convolutional layer followed by a linear output layer, and FiLM layers are included after each convolutional layer in $M_{task}$  \footnote{Implementation of this set of experiments is made available at: \url{https://github.com/google-research/google-research/tree/master/domain_conditional_predictors}}.

Results, as reported in Table \ref{tab:classification_results}, indicate that the conditioning approach boosts performance with respect to standard classifiers that disregard domain information as well as domain invariant approaches. In fact, the conditional predictors presented the highest out-of-domain accuracy amongst all evaluated methods. Surprisingly, in the ablation case where we drop $\mathcal{L}_{domain}$, domains can still be inferred from $z$ with high accuracy (c.f. Table \ref{tab:domain_classification}) which indicates the domain conditioning strategy enabled by the proposed architecture is exploited even if not enforced by an explicit training objective when multiple domains are present in the training sample.

\begin{table}[h]
\centering
\caption{Classification performance in terms of top-1 accuracy (\%). In-domain performance is measured using distorted MNIST test images while out-of-domain results correspond to evaluation on the standard test set of MNIST. The ablation with a learned embedding layer can only be used for in-domain predictions. For the in-domain evaluation, we loop over the test data 10 times to reduce the evaluation variance since each test example will be transformed differently each time.}
\resizebox{\textwidth}{!}{
\begin{tabular}{ccc}
\hline
\textbf{Model}                 & \textbf{\begin{tabular}[c]{@{}c@{}}In-domain\\ test accuracy (\%)\end{tabular}} & \textbf{\begin{tabular}[c]{@{}c@{}}Out-of-domain\\ test accuracy (\%)\end{tabular}} \\ \hline
Unconditional baseline         & 94.97                                                                           & 92.51                                                                               \\
Adversarial baseline           & 89.19                                                                           & 88.49                                                                               \\
Ablation: switching $M_{domain}$ for an embedding layer    & 92.56                                                                           & --                                                                                  \\
Ablation: dropping $\mathcal{L}_{domain}$ & 96.20                                                                           & 92.94                                                                               \\
Conditional predictor (\emph{Ours})              & 96.00                                                                           & 93.66                                                                               \\ \hline
\end{tabular}}
\label{tab:classification_results}
\end{table}

\begin{table}[h]
\centering
\caption{Domain classification top-1 accuracy (\%) measured for predictions of the underlying domain when the same transformations applied to train data are applied to test examples. In the ablation case, a linear classifier is trained on top of $z$ with the rest of the model frozen.}
\begin{tabular}{cc}
\hline
\textbf{Model}                 & \textbf{Test accuracy (\%)} \\ \hline
Adversarial baseline           & 84.22              \\
Ablation: dropping $\mathcal{L}_{domain}$ & 96.01              \\
Conditional predictor (\emph{Ours})              & 99.90              \\ \hline
\end{tabular}
\label{tab:domain_classification}
\end{table}

The reported proof-of-concept evaluation provided indications that domain-conditional predictors might indeed offer competitive performance under domain shift relative to both invariant approaches as well as standard classifiers. We thus proceed and evaluate to which extent such conclusions hold when larger scale widely adopted evaluation settings are considered. Moreover, validation experiments showed evidence suggesting that $z$ is indeed domain-dependent, and interestingly, that is the case even when $\mathcal{L}_{domain}$ is dropped, and performance in this case is fairly close to the proposed conditional model. As such, we further use the following evaluation to try and understand the contributions in observed performance improvements given by the architectural changes resulting from the inclusion of FiLM layers, and the actual influence of domain-conditioning in the final performance.

\section{Domain Generalization Evaluation}
\label{sec:complete_evaluation}

\subsection{Experiments description}
\label{sec:eval_description}

Seeking to confirm the findings in the proof-of-concept evaluation reported in section \ref{sec:experiments}, we now turn our attention to common benchmarks used for the domain generalization case under the leave-one-domain-out setting described previously. We specifically aim to answer the following questions:

\begin{enumerate}
    \item Are domain-conditional models able to generalize to data sources other than the ones observed during training? How do they compare against more common domain-invariant approaches?
    \item Can in-domain performance be used to reliably determine the best out-of-domain performers?
    \item Are conditioning variables indeed domain-dependent? What happens when domain supervision is dropped?
    \item What is the best combination of FiLM layers and regularization in order to enable effective conditioning?
\end{enumerate}

We first discuss differences between the planned and executed evaluations, and define a regularization procedure for FiLM layers that we found to be necessary in order to avoid overfitting. Question 1 listed above is addressed in section \ref{sec:eval_benchmarking} where standard classifiers, domain-invariant models (both our implementation and published results), and proposed domain-conditional predictors are compared. In section \ref{sec:stopping_criteria}, given that in realistic practical settings one does not have access to unseen domains and thus can only be able to perform cross-validation using in-domain data, we evaluate the usefulness of different criteria relying on in-domain validation data by comparing best performers under such criteria against the best out-of-domain performers, i.e. those obtained by direct evaluation in the test data of unseen domains. Ablations are reported in section \ref{sec:eval_ablations} where we evaluate \emph{self-modulated models}, i.e. those with the same architecture as conditional predictors but trained without domain supervision, and further check the resulting performance given by simplified FiLM layers where either scale or offset parameters are dropped. Finally, we check for domain-dependency in the conditioning variable $z$ in section \ref{sec:z_properties} where model-based dependency tests, as defined in appendix B, are carried out to verify whether domain-specific factors are encoded in learned representations.

\paragraph{A note on the final choice of evaluation data and baselines}

As described in section \ref{sec:evaluation}, a subset of four candidate benchmarks would be employed for comparing different approaches. We then briefly describe the features of those datasets that guided the decision regarding which ones to actually use. We found VLCS \citep{fang2013unbiased}, for instance, to be relatively small in terms of training sample size when considering the size of models under analysis. Moreover, one of the domains within VLCS, Caltech-101, is a potential subset of ImageNet, which is commonly used to pre-train models for domain generalization. Yet another relevant factor to consider is the fact that VLCS has all 4 domains corresponding to natural images, representing not much of a shift across training and testing domains. For the case of OfficeHome \citep{venkateswara2017Deep}, we found previously reported results  to be high enough to leave little room for improvement (c.f. \citep{peng2019moment} for the domain adaptation case, for instance), and, similarly to VLCS, it is not very diverse in terms of domains but also in terms of the classes considered since they all correspond to office objects. We thus focus our resource budget on: PACS \citep{li2017deeper} for the smaller scale range (but large enough to avoid instabilities) in terms of both sample size and number and diversity of classes/domains, and DomainNet \citep{peng2019moment} for a much larger scale case. In terms of baselines, we found performances reported across IRM \citep{arjovsky2019invariant} and Rex \citep{krueger2020out} to not be significantly different and then pick only one of them. We remark both approaches attempt to train invariant classifiers for varying domains. Moreover, we included extra baselines for a complete comparison. For the case of PACS, the domain-invariant approach reported by \cite{zhao2020domain} is considered, while challenging privileged baselines that have access to unlabeled data from the target domain were also included for the case of DomainNet; those correspond to \citep{saito2018maximum,xu2018deep,peng2019moment}.

\subsection{Regularized FiLM}
\label{sec:film_regularization}

Given the extra capacity provided by the inclusion of FiLM layers, we found it to be necessary to include regularization strategies aimed at avoiding conditional models of growing too complex, which could incur in overfitting. We thus consider penalizing FiLM layers when they differ from the \emph{identity map}\footnote{The identity map outputs its inputs.}, and do so through the mean squared error (MSE) measured between inputs and outputs of FiLM layers throughout the model. The regularization penalty, indicated by $\Omega_{FiLM}$, will be given by the average MSE given by:

\begin{equation}
    \label{eq:film_reg}
    \Omega_{FiLM} = \frac{1}{|\mathcal{F}|} \sum_{k \in \mathcal{F}} \text{MSE}(x^{k-1}, FiLM^k(x^{k-1}, z))
\end{equation}
where $\mathcal{F}$ stands for the set of indexes of FiLM layers while $k$ indicates general indexes of layers within $M_{task}$. The inputs to the conditioning layers correspond to $z$, the external conditioning variable, and $x^{k-1}$, representing the output of the layer preceding $FiLM^k$. The complete training loss then becomes:

\begin{equation}
\label{eq:full_training_objective}
\mathcal{L}=(1-\lambda)\mathcal{L}_{task}+\lambda\mathcal{L}_{domain}+\gamma\Omega_{FiLM},
\end{equation}
where $\gamma$ is a hyperparameter weighing the importance of the penalty in the overall training objective, which allows controlling how far from the identity operator FiLM layers are able to reach. We remark that the case where $\gamma=0$ is included within the range of all of our hyperparameter search so that the standard unconstrained FiLM can always be selected by the search procedure. We, however, found that adding some regularization consistently yielded better generalization, but $\gamma$ was usually assigned to relatively small values such as $10^{-9}$ or $10^{-10}$.

\subsection{Domain-conditional predictors are much simpler than and as performant as domain-invariant approaches}
\label{sec:eval_benchmarking}

In the following, unless stated otherwise, results are reported in terms of 95\% confidence intervals of the top-1 accuracy obtained with a sample of size 5 (i.e. 5 independent training runs were executed to generate each entry in the tables). Across tables, columns indicate left-out domains that were completely disregarded to train the models whose performance is reported under said column. Our models as well as most baselines correspond to the ResNet-50 architecture, and we follow the common practice of pre-training the models on ImageNet. We also freeze batch normalization layers during fine-tuning to comply with previously reported evaluation protocols for the considered tasks. Four FiLM layers are included in $M_{task}$ in total corresponding to one after each ResNet block. Further implementation details are included in the appendix.

\subsubsection{Matching the performance of state-of-the-art invariant approaches on PACS}

Evaluation for PACS is reported in Table \ref{tab:pacs_results}, and two distinct cases are considered consisting of \emph{in-domain performance}, which indicates the evaluation is carried out using fresh data (unseen during training) from the same domains used for training, while the \emph{out-of-domain} case consists of the evaluation performed on the test partition of left-out domains. Moreover, in order to enable a fair comparison with published results of competing approaches, out-of-domain performance is reported in terms of the best accuracy obtained by a given model throughout training, which we indicate by the term \emph{Oracle} (c.f. section \ref{sec:stopping_criteria} for a detailed discussion on different methods for selecting models for evaluation). Performance is reported for the conditional models proposed here along with a number of baselines. We use the term \emph{unconditional models} to refer to models trained under standard ERM and that have the same architecture as $M_{task}$ after removing conditioning layers, i.e. a standard ResNet-50. In order to control for the effect in performance given by the differences in model sizes, we further considered higher capacity versions of the unconditional baseline, where we doubled the width of the model by using two models trained side-by-side and concatenating their outputs prior to the final layer, and also included a deeper model implemented as a ResNet-101. For the adversarial case as well as IRM, we used the exact same model as ours corresponding to the pair $M_{task}$ and $M_{domain}$ as well as FiLM layers, and no domain-related supervision is performed in such cases. State-of-the-art results achieved by the domain-invariant approach reported by \cite{zhao2020domain} are finally included.

We highlight that, in the out-of-domain case and across most domains, the conditional model is able to outperform unconditional baselines regardless of their size as well as the domain-invariant approaches that we implemented, and does so without affecting its in-domain performance, which did occur for the domain-invariant baselines. This suggests that inducing invariance has a cost in terms of generalization performance for training domains which was not observed for conditional approaches. Moreover, in-domain performance seems to be a good predictor of out-of-domain differences across models for most cases. As will be further discussed later on, that doesn't appear to be always the case when using in-domain performance to do model selection or deciding when to stop training. We further remark that the conditional approach is able to closely match the performance of the state-of-the-art domain-invariant scheme while doing so with practical advantages such as a much simpler training scheme and less hyperparameters to be tuned. We further mention that, as observed in previous work \citep{gulrajani2020search}, unconditional models trained under standard ERM perform surprisingly well in the domain generalization case, and the reason for that remains an open question\footnote{Considering that the domain generalization setting violates assumptions required in ERM in order to guarantee generalization.}. However, in section \ref{sec:z_properties}, we show representations learned through ERM are domain-dependent for the types of data we consider. Regarding some of the invariant baselines we highlight that, similarly to what has been reported in the past \citep{albuquerque2019generalizing,krueger2020out,rosenfeld2020risks}, we found IRM to perform similarly to ERM (referred to as unconditional in the tables). Moreover, considering the case of the adversarial baseline, its performance gap with respect to other approaches is likely due to the added complexity resulting from the adversarial training scheme and the number of hyperparameters it incurs in addition to those conditional models bring in. We thus claim the overall simplicity of the domain-conditional setting as a relevant practical advantage over adversarial formulations of invariant approaches.

\begin{table}[]
\centering
\caption{Leave-one-domain-out evaluation on PACS. Results correspond to 95\% confidence intervals on the top-1 prediction accuracy (\%). Results in bold indicate the best performers for each left-out domain. Conditional models perform on pair with a state-of-the-art invariant approach while being much simpler to train. 
\label{tab:pacs_results}}
\begin{tabular}{ccccc}
\hline
\multicolumn{1}{l}{} & \multicolumn{4}{c}{\textbf{Left-out-domain}}                                                                    \\ \cline{2-5} 
                     & \textit{\textbf{Art Painting}} & \textit{\textbf{Cartoon}} & \textit{\textbf{Photo}} & \textit{\textbf{Sketch}} \\ \hline
\multicolumn{5}{l}{\textit{In-domain test accuracy}}                                                                                            \\ \hline
Unconditional        & 95.48$\pm$0.31                    & 94.52$\pm$0.18               & 94.26$\pm$0.28             & 95.91$\pm$0.24              \\
Uncond. (\emph{Wide})       & \textbf{96.20$\pm$0.29}                    & 94.62$\pm$0.33               & 94.42$\pm$0.39             & 96.17$\pm$0.19              \\
Uncond. (\emph{Deep})       & 95.83$\pm$0.24                    & 94.41$\pm$0.25               & 94.50$\pm$0.19             & 95.19$\pm$0.32              \\
Adversarial          & 91.96$\pm$0.55                    & 91.97$\pm$0.76               & 90.60$\pm$0.75             & 93.21$\pm$0.45              \\
IRM                  & 94.99$\pm$0.50                    & 94.00$\pm$0.18               & 93.26$\pm$0.30             & 95.29$\pm$0.32              \\
Conditional (\emph{Ours})          & 95.86$\pm$0.31                    & \textbf{95.06$\pm$0.15}               & \textbf{94.90$\pm$0.49}             & \textbf{96.33$\pm$0.29}              \\ \hline
\multicolumn{5}{l}{\textit{Oracle out-of-domain test accuracy}}                                                                        \\ \hline
Unconditional        & 85.40$\pm$0.57                    & 77.28$\pm$0.78               & 95.23$\pm$0.53             & 74.58$\pm$1.09              \\
Uncond. (\emph{Wide})       & 85.98$\pm$0.45                    & 78.37$\pm$0.66               & 95.53$\pm$0.44             & 74.31$\pm$1.11              \\
Uncond. (\emph{Deep})       & 84.81$\pm$0.54                    & 79.47$\pm$0.51               & 95.09$\pm$0.21             & 76.67$\pm$2.36              \\
Adversarial          & 74.74$\pm$1.44                    & 74.64$\pm$0.81               & 92.67$\pm$1.03             & \textbf{78.65$\pm$1.08}              \\
IRM                  & 81.73$\pm$0.56                    & 76.53$\pm$0.57               & 95.07$\pm$0.29             & 77.76$\pm$2.18              \\
Conditional (\emph{Ours})          & 86.28$\pm$0.84                    & \textbf{80.29$\pm$1.48}               & 96.75$\pm$0.54             & 77.34$\pm$2.16              \\ \hline
\cite{zhao2020domain}              & \textbf{87.51$\pm$1.03}                    & 79.31$\pm$1.40               & \textbf{98.25$\pm$0.12}             & 76.30$\pm$0.65              \\ \hline
\end{tabular}
\end{table}

\subsubsection{Closing the gap between models trained with access to unlabeled target data and domain generalization approaches on DomainNet}

For the case of DomainNet, results, as reported in Table \ref{tab:domainnet_results}, include what we refer to as \emph{privileged baselines}, indicating that such models have an advantage in that they have access to the test distribution through an unlabeled data sample made available to them during training\footnote{The setting under consideration in the case of privileged baselines is commonly referred to as multi-source unsupervised domain adaptation.}. Moreover, the three considered privileged baselines are implemented using deeper models corresponding to a ResNet-101 architecture as discussed by \cite{peng2019moment}, rendering the comparison unfair in their favor. We then compare such approaches with conditional models as well as the unconditional cases when trained under the domain generalization setting, i.e. with no access to the test domain during training. Once more, conditional models are observed to outperform standard classifiers even if the overall number of parameters is roughly matched considering the wide/deep cases. More importantly, conditional models are observed to be able to perform on pair with baselines that have access to the test domain, showing the proposed approach to reduce the gap between models that specialize to a particular target domain and those that are simply trained on diverse data sources, without focusing on any test distribution in particular. This is of practical significance since it can enable deployment of models that readily generalize to new sources, without requiring any prior data collection.

\begin{table}[]
\centering
\caption{Leave-one-domain-out evaluation on DomainNet. Results correspond to 95\% confidence intervals on the top-1 prediction accuracy  (\%). Results in bold indicate the best performers for each left-out domain. In all cases, conditional predictors outperform at least one of the privileged baselines. 
\label{tab:domainnet_results}}
\resizebox{\textwidth}{!}{
\begin{tabular}{ccccccc}
\hline
                   & \multicolumn{6}{c}{\textbf{Left-out domain}}                                                                     \\ \cline{2-7} 
                   & \textbf{Clipart} & \textbf{Infograph} & \textbf{Painting} & \textbf{Quickdraw} & \textbf{Real} & \textbf{Sketch} \\ \hline
\multicolumn{7}{l}{\textit{In-domain test accuracy}}                                                                                           \\ \hline
Unconditional      & 60.85$\pm$0.78      & 64.42$\pm$1.26        & 62.29$\pm$1.47       & 62.07$\pm$1.19        & 57.19$\pm$1.66   & 62.30$\pm$1.61     \\
Uncond. (\emph{Wide})     & 61.67$\pm$1.85      & 63.99$\pm$1.48        & 63.18$\pm$1.13       & 63.70$\pm$0.92        & 58.37$\pm$0.64   & 63.98$\pm$0.99     \\
Uncond. (\emph{Deep})     & 61.22$\pm$0.94      & 65.01$\pm$1.86        & 63.08$\pm$1.40       & 62.88$\pm$0.96        & 57.88$\pm$0.69   & 63.48$\pm$1.08     \\
Conditional (\emph{Ours}) & \textbf{65.10$\pm$0.47}      & \textbf{69.01$\pm$0.75}        & \textbf{66.99$\pm$0.49}       & \textbf{67.74$\pm$0.44}        & \textbf{60.90$\pm$0.59}   & \textbf{67.43$\pm$0.48}     \\ \hline
\multicolumn{7}{l}{\textit{Oracle out-of-domain test accuracy}}                                                                       \\ \hline
Unconditional      & 52.47$\pm$1.13      & 19.64$\pm$0.42        & 44.12$\pm$0.71       & 11.67$\pm$0.50        & 51.94$\pm$1.68   & 44.40$\pm$1.42     \\
Uncond. (\emph{Wide})     & 54.12$\pm$2.25      & 18.96$\pm$0.62        & 44.88$\pm$0.83       & 12.16$\pm$0.16        & 52.03$\pm$1.12   & 45.51$\pm$1.68     \\
Uncond. (\emph{Deep})     & 54.70$\pm$1.26      & 21.05$\pm$0.44        & 45.54$\pm$0.66       & 12.71$\pm$0.33        & 51.62$\pm$1.43   & 46.83$\pm$1.17     \\
Conditional (\emph{Ours}) & 58.37$\pm$0.67      & 23.25$\pm$0.45        & 50.06$\pm$0.43       & \textbf{13.32$\pm$0.34}        & 57.25$\pm$0.47   & \textbf{50.52$\pm$1.05}     \\ \hline
\multicolumn{7}{l}{\textit{Privileged baselines with access to unlabeled target data}}                                                \\ \hline
\cite{saito2018maximum}           & 54.3$\pm$0.64       & 22.1$\pm$0.70         & 45.7$\pm$0.63        & 7.6$\pm$0.49          & 58.4$\pm$0.65    & 43.5$\pm$0.57      \\
\cite{xu2018deep}           & 48.6$\pm$0.73       & 23.5$\pm$0.59         & 48.8$\pm$0.63        & 7.2$\pm$0.46          & 53.5$\pm$0.56    & 47.3$\pm$0.47      \\
\cite{peng2019moment}           & \textbf{58.6$\pm$0.53}       & \textbf{26.0$\pm$0.89}         & \textbf{52.3$\pm$0.55}        & 6.3$\pm$0.58          & \textbf{62.7$\pm$0.51}    & 49.5$\pm$0.76      \\ \hline
\end{tabular}
}
\end{table}

\subsection{In-domain performance is a better predictor of out-of-domain accuracy for larger scale data}
\label{sec:stopping_criteria}

Following the discussion in previous work \citep{albuquerque2019generalizing,krueger2020out}, we study the impact in out-of-domain performance resulting from the choice of different criteria used to decide when to stop training. This is an important analysis which we argue should be reported by any work proposing approaches targeting domain generalization since performing cross-validation in this setting is non-trivial, given that no specific test domain is defined at training time. We thus consider the performance of the proposed models when two different criteria relying solely on in-domain validation data are used, and compare them against the case where the test data is used to select the best performing model (i.e. the \emph{oracle} case). In further detail, we follow \cite{albuquerque2019generalizing} and analyze the out-of-domain performance obtained by models that reached the highest in-domain prediction accuracy and the lowest in-domain classification loss, both measured with fresh in-domain data held out from training.

\begin{table}[]
\centering
\caption{Comparison of different stopping criteria on PACS. Results in terms of out-of-domain accuracy (\%) are shown for each left-out-domain and each method for selecting models to be evaluated. The best in-domain models are significantly worse than the best out-of-domain performers.
\label{tab:pacs_stopping}}
\begin{tabular}{ccccc}
\hline
\multicolumn{1}{l}{} & \multicolumn{4}{c}{\textbf{Left-out domain}}                                                          \\ \cline{2-5} 
                     & \textbf{Art painting}   & \textbf{Cartoon}        & \textbf{Photo}          & \textbf{Sketch}         \\ \hline
\multicolumn{5}{l}{\textit{Best in-domain performers evaluated out-of-domain}}                                                        \\ \hline
Validation accuracy  & 80.65$\pm$1.83          & 75.80$\pm$2.81          & 95.34$\pm$0.62          & 71.76$\pm$2.60          \\
Validation loss      & 83.84$\pm$0.84          & 74.90$\pm$2.79          & 95.80$\pm$0.67          & 72.59$\pm$2.34          \\ \hline
\multicolumn{5}{l}{\textit{Best out-of-domain performer}}                                                                    \\ \hline
Oracle               & \textbf{86.28$\pm$0.84} & \textbf{80.29$\pm$1.48} & \textbf{96.75$\pm$0.54} & \textbf{77.34$\pm$2.16} \\ \hline
\end{tabular}
\end{table}

Results are reported in Tables \ref{tab:pacs_stopping} and \ref{tab:domainnet_stopping} for the cases of PACS and DomainNet, respectively. For the case of PACS, we observe a significant gap across all domains between the best performance which can be achieved and that observed by the best in-domain performer. The gap reduces drastically when we move to the larger scale case corresponding to DomainNet. We attribute the smaller gap in the case of DomainNet to the increased domain diversity since more distinct training sources are available in that case. Considering both datasets, we did not observe a significant difference in selecting either of the considered practical stopping criteria that only use in-domain validation data.

\begin{table}[]
\centering
\caption{Comparison of different stopping criteria on DomainNet. Results in terms of out-of-domain accuracy (\%) are shown for each left-out-domain and each method for selecting models to be evaluated. A smaller gap is observed for DomainNet between model selection approaches that use in-domain validation data and those that have access to the test set of the left-out-domain.
\label{tab:domainnet_stopping}}
\resizebox{\textwidth}{!}{
\begin{tabular}{ccccccc}
\hline
\multicolumn{1}{l}{} & \multicolumn{6}{c}{\textbf{Left-out domain}}                                                                                                              \\ \cline{2-7} 
                     & \textbf{Clipart}        & \textbf{Infograph}      & \textbf{Painting}       & \textbf{Quickdraw}      & \textbf{Real}           & \textbf{Sketch}         \\ \hline
\multicolumn{7}{l}{\textit{Best in-domain performers evaluated out-of-domain}}                                                                                                   \\ \hline
Validation accuracy  & 57.56$\pm$0.77          & 22.30$\pm$0.44          & 49.51$\pm$0.61          & 12.35$\pm$0.57          & 56.32$\pm$0.28          & 49.89$\pm$1.12          \\
Validation loss      & 58.04$\pm$0.99          & 22.56$\pm$0.57          & 49.33$\pm$0.61          & 12.58$\pm$0.62          & 56.45$\pm$1.30          & 49.95$\pm$1.40          \\ \hline
\multicolumn{7}{l}{\textit{Best out-of-domain performer}}                                                                                                                        \\ \hline
Oracle               & \textbf{58.37$\pm$0.67} & \textbf{23.25$\pm$0.45} & \textbf{50.06$\pm$0.43} & \textbf{13.32$\pm$0.34} & \textbf{57.25$\pm$0.47} & \textbf{50.52$\pm$1.05} \\ \hline
\end{tabular}
}
\end{table}

\subsection{Ablations}
\label{sec:eval_ablations}

\subsubsection{The self-modulated case: dropping domain supervision does not affect performance}

We now perform ablations in order to understand what are the sources of improvements provided by the discussed conditioning mechanism. Specifically, we drop the domain supervision term of the training objective defined in eq. \ref{eq:full_training_objective} to check whether the improvements observed in results reported in Tables \ref{tab:pacs_results} and \ref{tab:domainnet_results} with respect to unconditional and invariant baselines are due to the increase in model size rather than the conditioning approach. We then retrain our models while setting $\lambda=0$, in which case we refer to the resulting model as \emph{self-modulated}, given that in this case there's no supervision signal indicating what kind of information should be encoded in $z=M_{domain}(x)$ other than the classification criterion $\mathcal{L}_{task}$. Results are reported in Tables \ref{tab:pacs_selfmod} and \ref{tab:domainnet_selfmod} for the cases of PACS and DomainNet, respectively. In both datasets and for all left-out domains, we do not observe significant performance differences across the conditional and self-modulated models. We thus claim one of the two following statements explain the matching performance across the two schemes:

\begin{enumerate}
    \item Even if domain supervision is not employed, $z$ is still domain-dependent.
    \item Conditioning on domain information does not improve out-of-distribution performance, and improvements with respect to unconditional models are simply due to architectural changes.
\end{enumerate}

We provide evidence in section \ref{sec:z_properties} supporting the first statement enumerated on the above, where we show linear classifiers to be able to discriminate domains using only $z$ obtained after training self-modulated models with relatively high accuracy. Such result indicates that domain-dependent factors will be encoded in $z$ even if this property \emph{is not} explicitly enforced via supervision. This finding is of practical relevance given that no domain labels are required to enable domain-conditioning in this type of architecture, yielding models that perform better than standard classifiers and on pair with complex domain-invariant approaches that do require domain labels. We remark, however, that the conditional setting offers the advantage of enabling domain-related inferences, which are not supported in the self-modulated case. Given that both cases present a similar performance, we argue that the choice regarding which setting to use in a practical application should be guided by factors such as the availability of domain labels as well as the need for performing domain predictions at testing time.

\begin{table}[]
\centering
\caption{Ablation: self-modulated models compared with domain-conditional predictors on PACS. Results correspond to the best out-of-domain prediction accuracy (\%) obtained by each approach. Best results for each domain are in bold. No significant performance difference is observed across approaches.
\label{tab:pacs_selfmod}}
\begin{tabular}{ccccc}
\hline
               & \multicolumn{4}{c}{\textbf{Left-out domain}}                                \\ \cline{2-5} 
               & \textbf{Art painting} & \textbf{Cartoon} & \textbf{Photo} & \textbf{Sketch} \\ \hline
Conditional    & \textbf{86.28$\pm$0.84}           & \textbf{80.29$\pm$1.48}      & \textbf{96.75$\pm$0.54}    & 77.34$\pm$2.16     \\
Self-modulated & 85.63$\pm$0.58           & 80.21$\pm$0.60      & 96.42$\pm$0.19    & \textbf{78.80$\pm$1.31}     \\ \hline
\end{tabular}
\end{table}

\begin{table}[]
\centering
\caption{Self-modulated models compared with domain-conditional predictors on DomainNet. Results correspond to the best out-of-domain prediction accuracy (\%) obtained by each approach. Best results for each domain are in bold.  No significant performance difference is observed across approaches.
\label{tab:domainnet_selfmod}}
\resizebox{\textwidth}{!}{
\begin{tabular}{ccccccc}
\hline
               & \multicolumn{6}{c}{\textbf{Left-out domain}}                                                                     \\ \cline{2-7} 
               & \textbf{Clipart} & \textbf{Infograph} & \textbf{Painting} & \textbf{Quickdraw} & \textbf{Real} & \textbf{Sketch} \\ \hline
Conditional    & \textbf{58.37$\pm$0.67}      & \textbf{23.25$\pm$0.45}        & 50.06$\pm$0.43       & \textbf{13.32$\pm$0.34}        & 57.25$\pm$0.47   & \textbf{50.52$\pm$1.05}     \\
Self-modulated & 58.35$\pm$0.94      & 22.86$\pm$0.26        & \textbf{50.31$\pm$0.45}       & 12.87$\pm$0.21        & \textbf{57.69$\pm$0.70}   & 50.01$\pm$0.61     \\ \hline
\end{tabular}
}
\end{table}

\begin{figure}[h]
\centering
\begin{subfigure}{\textwidth}
  \centering
  \includegraphics[width=.65\linewidth]{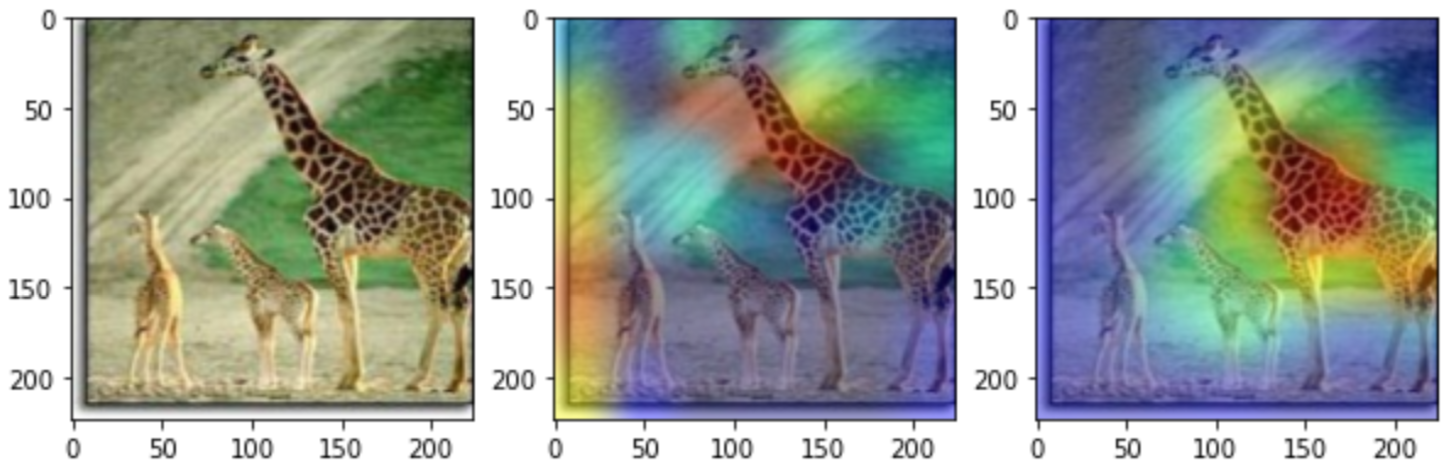}
  \caption{Grad-CAM heat-maps resulting from the \emph{domain-conditional model}. Images correspond to the original input, and the results of the analysis performed using $M_{domain}$ and $M_{task}$, respectively. While $M_{task}$ focus on the object of interest, $M_{domain}$ accounts for secondary features including the background.}
  \label{fig:gradcam_conditional}
\end{subfigure}

\begin{subfigure}{\textwidth}
  \centering
  \includegraphics[width=.65\linewidth]{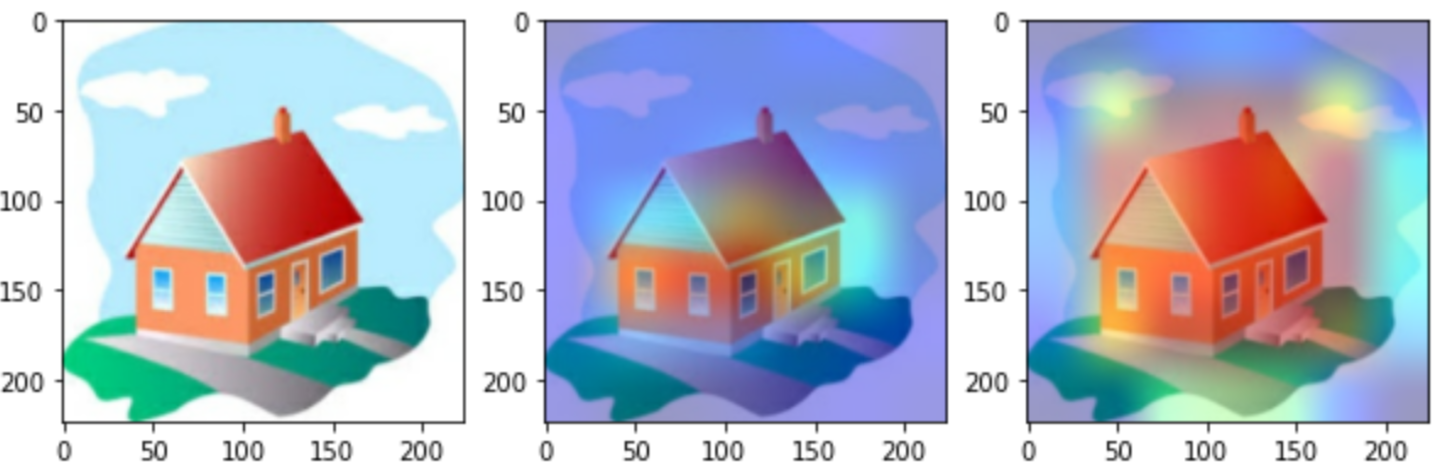}
  \caption{Grad-CAM heat-maps resulting from \emph{self-modulated model}. Images correspond to the original input, and the results of the analysis performed using $M_{domain}$ and $M_{task}$, respectively. In this case, $M_{task}$ spreads its focus over most of the input image while $M_{domain}$ targets specific regions.}
  \label{fig:gradcam_selfmodulated}
\end{subfigure}
\caption{Heat-maps indicating parts of input images implying predictions for the different models we consider.}
\end{figure}

We further qualitatively assess the differences in behavior between the two approaches using Grad-CAM heat-maps \citep{selvaraju2017grad}, as shown in Figures \ref{fig:gradcam_conditional} and \ref{fig:gradcam_selfmodulated} for the cases of conditional and self-modulated models, respectively. Those indicate parts of the input deemed relevant to resulting predictions, and in both cases, models employed in the analysis were trained when \emph{sketch} was arbitrarily chosen as the left-out-domain. From the left to the right, each group of three images corresponds to the original image, and Grad-CAM plots obtained from $M_{domain}$ and $M_{task}$, respectively. One can then observe that, for the case of the conditional model, $M_{domain}$ accounts for regions of the input other than the object of interest such as the background. On the other hand, $M_{task}$ attends to more specific parts of the input that lie on the foreground. Interestingly, this behavior seems to be reverted when the self-modulated case is evaluated. In this case, while $M_{domain}$ concentrates in specific regions of the object defining the underlying class, $M_{task}$ accounts for the entire input. In both cases however, the two models learn to work in a complementary fashion in such a way that $M_{task}$ and $M_{domain}$ often focus on different aspects of the input data. Further differences between the two schemes will be discussed in section \ref{sec:z_properties}.

\subsubsection{Dropping FiLM parameters: Full FiLM layers along with regularization yield the best conditioning scheme}

We now perform a second set of ablations to determine whether simpler variations of FiLM layers are able to perform similarly to the full model. In table \ref{tab:pacs_filmablation}, results considering PACS are reported, and the full conditional model is compared against its variations where either the scale or the offset parameters are dropped from all FiLM layers throughout the model. Results indicate that using full FiLM yields better performance across most of the domains. We further highlight that $\gamma$, the parameter controlling the contribution of the regularization penalty $\Omega_{FiLM}$, was consistently selected to non-zero values. We then conclude that using the full FiLM layers and employing the regularization strategy described in eq. \ref{eq:film_reg} is the best performing combination for the types of models/problems we consider. We finally remark that the combination of over-parameterization and regularization is a common approach within recent work in neural networks, which is aligned with the setting we observed to yield the best performing predictors.

\begin{table}[]
\centering
\caption{Dropping parameters of FiLM layers. Results are reported for PACS in terms of the best out-of-domain prediction accuracy (\%) obtained by each approach. Best results for each domain are in bold.
\label{tab:pacs_filmablation}}
\begin{tabular}{ccccc}
\hline
            & \multicolumn{4}{c}{\textbf{Left-out domain}}                                \\ \cline{2-5} 
            & \textbf{Art painting} & \textbf{Cartoon} & \textbf{Photo} & \textbf{Sketch} \\ \hline
Conditional & \textbf{86.28$\pm$0.84}           & \textbf{80.29$\pm$1.48}      & \textbf{96.75$\pm$0.54}    & 77.34$\pm$2.16     \\ \hline
- scale     & 84.79$\pm$0.61           & 78.68$\pm$1.34      & 96.50$\pm$0.10    & 75.96$\pm$2.11     \\
- offset    & 85.09$\pm$0.28           & 78.68$\pm$0.80      & 95.86$\pm$0.37    & \textbf{77.46$\pm$0.99}     \\ \hline
\end{tabular}
\end{table}

\subsection{Understanding properties of $z$: domain-dependency is observed with or without domain supervision}
\label{sec:z_properties}

In the following set of experiments, we study whether domain-dependency is indeed achieved in $z=M_{domain}(x)$, and further investigate what properties appear in $z$ when the self-modulated case is considered. Representations output by unconditional models are also evaluated in order to help understanding the properties of representations resulting from ERM with standard architectures. In order to perform such evaluation, we carry out a dependency analysis between representations extracted from the models and domain or class labels\footnote{Refer to the supplementary materials for a definition of the model-based statistical dependency analysis performed in this set of experiments.}. We specifically train two linear classifiers using the domain representations $z$ for both the cases of conditional and self-modulated models, and further considered the output of the convolutional stack of the unconditional ResNet-50. While one of the classifiers is trained to predict the domains, the other predicts the actual class labels, and these steps are repeated for the test data of training domains for representations obtained by the models trained when each domain is left-out. Extra results corresponding to low-rank projections of $z$ as well as a similar analysis performed using FiLM parameters rather than $z$ are included in the appendix.

Results reported in Tables \ref{tab:pacs_domainclassification} and \ref{tab:pacs_taskclassification} correspond to 95\% confidence intervals of the in-domain test accuracy obtained through 5-fold cross-validation given by logistic regression to predict domains or classes, respectively. Considering the conditional model, representations are clearly domain-dependent as originally expected, and class prediction is not as accurate as in the self-modulated case. Interestingly, the embedding spaces given by the self-modulated models are both domain- and class-dependent, and remarkably, one can predict classes from them with an accuracy comparable to the actual classifier (compare to the in-domain case in table \ref{tab:pacs_results}), while domains can be predicted with high accuracy relative to a random predictor in all of the considered cases as well.

\begin{table}[h]
\centering
\caption{Prediction accuracy (\%) of logistic regression for performing domain classification on top of $z$. Results correspond to 95\% confidence intervals obtained through 5-fold cross validation.
\label{tab:pacs_domainclassification}}
\begin{tabular}{ccccc}
\hline
               & \multicolumn{4}{c}{\textbf{Left-out domain}}                                \\ \cline{2-5} 
               & \textbf{Art painting} & \textbf{Cartoon} & \textbf{Photo} & \textbf{Sketch} \\ \hline
Conditional    & 98.88$\pm$0.18           & 98.19$\pm$0.61      & 99.76$\pm$0.11    & 96.10$\pm$0.51     \\
Self-modulated & 72.45$\pm$1.32           & 80.95$\pm$0.82      & 90.86$\pm$0.69    & 62.02$\pm$0.91     \\ \hline
Unconditional  & 98.14$\pm$0.31           & 95.11$\pm$0.47      & 98.34$\pm$0.17    & 90.43$\pm$1.16     \\ \hline
\end{tabular}
\end{table}

\begin{table}[h]
\centering
\caption{Prediction accuracy (\%) of logistic regression for predicting class labels using $z$. Results correspond to 95\% confidence intervals obtained through 5-fold cross validation.
\label{tab:pacs_taskclassification}}
\begin{tabular}{ccccc}
\hline
               & \multicolumn{4}{c}{\textbf{Left-out domain}}                                \\ \cline{2-5} 
               & \textbf{Art painting} & \textbf{Cartoon} & \textbf{Photo} & \textbf{Sketch} \\ \hline
Conditional    & 79.65$\pm$1.30           & 76.71$\pm$0.42      & 66.55$\pm$1.30    & 73.71$\pm$1.07     \\
Self-modulated & 93.05$\pm$0.32           & 92.14$\pm$0.96      & 90.87$\pm$0.61    & 91.72$\pm$0.74     \\ \hline
Unconditional  & 94.67$\pm$0.67           & 94.47$\pm$0.67      & 94.90$\pm$0.82    & 95.78$\pm$0.51     \\ \hline
\end{tabular}
\end{table}

Such observations support the claim that domain-conditional modeling is indeed being enforced by the proposed architecture, and, most notably, this is the case even if domain supervision is dropped. For the unconditional case, it was recently pointed out that simple ERM makes it for strong baselines in the domain generalization setting \citep{gulrajani2020search}. We found that representations learned by this type of model, in addition to being class-dependent as expected given that those features are obtained prior to the output layer, they are also domain-dependent in that domains can be predicted from the features with high accuracy by linear classifiers. This suggests that domain-dependent high level features are accounted for in this case, and this observation serves to both explain why ERM somewhat works out-of-domain, as well as to motivate domain-conditional approaches, further considering that including explicit conditioning mechanisms in the architecture improved performance relative to unconditional models.

\section{Conclusion}
\label{sec:conclusion}

In the following, we address the questions posed in section \ref{sec:eval_description} and additionally discuss further conclusions drawn from the reported evaluation.

\paragraph{Which approach is finally recommended? Domain-invariant or domain-conditional models?}

We found both approaches to work similarly well in terms of final out-of-domain performance. We however argue that the domain-conditional approach yields several practical advantages such as: (\textbf{i})-overall simplicity in that no adversarial training is required and a standard maximum likelihood estimation scheme is employed; (\textbf{ii})-Less hyperparameters need to be tuned for the domain-conditional case; (\textbf{iii})-There is no reliance on domain labels since the self-modulating case generates domain-dependent representations with no domain supervision.

\paragraph{Which stopping criterion should be used for domain generalization?}

In practice, one might not have any access to data sources over which models are expected to work well, and thus deciding when to stop training and which version of models to use at testing time becomes tricky. In our evaluation, we did not observe significant differences in using either in-domain loss or accuracy as a stopping criterion. However, we found that the scale of the data in terms of both domain diversity and sample size helps in reducing the gap between best in-domain and out-of-domain selected models. A practical recommendation is then to ensure datasets are created including diverse sets of data sources.

\paragraph{Are domain labels necessary for achieving out-of-distribution generalization?}

We found that dropping domain supervision does not affect performance as much as originally expected. This is due to the fact that learned conditioning representations $z$ are still domain-dependent (in addition to being strongly class-dependent) even if such property is not enforced via supervision, not affecting the conditioning mechanism significantly. We thus claim that the main requirements for achieving out-of-distribution generalization under the domain-conditional framework, besides having explicit conditioning mechanisms built in the architecture, is to have diverse enough training data in the sense that distinct data sources are available. In summary, if no domain-related inference is required for the application of interest, domain supervision can be dropped without significantly affecting performance.

\paragraph{Why does ERM achieve a relatively high accuracy in unseen data sources?}

We found evidence showing that domain-dependent representations are learned by ERM in the sense that domains can be easily inferred from representations it yields. We then argue that this behaviour suggests that a similar modeling approach to the one we propose herein naturally occurs when performing ERM, given a model family with enough capacity. While this is an interesting feature of ERM, there's just so much one can do without explicitly enforcing the conditioning behaviour. For instance, we observed that increasing model size did not result in better performance. On the other hand, introducing explicit conditioning mechanisms in the architectures does enable improvements in out-of-domain performance.

\paragraph{What are some open questions and directions for domain generalization research under the domain-conditional setting?}

Computing domain representations $z$ in the proposed setting is relatively expensive in that a separate model is used for that purpose. Studying approaches alleviating that cost by either reducing the size of $M_{domain}$ or dropping it altogether constitutes a natural extension of the method discussed herein. In addition to that, extending the conditional setting to cases where some information is available regarding possible test data sources is a relevant direction given its practical applicability. This could be achieved by either leveraging an unlabeled data sample from a particular target distribution, or using available metadata to infer, for instance, which training domain is closest to a target of interest, and focusing on that particular domain at training time.

\bibliography{bibliography.bib}

\clearpage
\section*{Supplementary material}

\subsection*{Appendix A: Training details}

Most of evaluated models correspond to a ResNet-50, except for the case of unconditional models where pairs of ResNet-50 or a ResNet-101 is considered (those cases are indicated in the tables by ``wide'' or ``deep'', respectively). Parameters updates are carried out according to Adam \citep{kingma2014adam} without any external schedule for decaying the learning rate. The training objective described in eq. \ref{eq:full_training_objective} is employed, and further regularization strategies are considered corresponding to a $L_2$ penalty on the norm of the parameters of $M_{task}$ and $M_{domain}$ (FiLM layers are not considered since $\Omega$ accounts for bounding the complexity of those layers) as well as label smoothing applied only to compute $\mathcal{L}_{task}$. Training hyperparameters are shown in Table \ref{tab:hyperparameters}. For the self-modulated case, we reuse all hyperparameters and set $\lambda=0$.

\begin{table}[h]
\centering
\caption{Hyperparameters used to train models corresponding to results reported across the evaluation cases. The self-modulated case corresponds to setting $\lambda=0$.
\label{tab:hyperparameters}}
\begin{tabular}{ccc}
\hline
\multirow{2}{*}{\textbf{Hyperparameter}} & \multicolumn{2}{c}{\textbf{Dataset}}     \\ \cline{2-3} 
                                & \textbf{PACS}      & \textbf{DomainNet}  \\ \hline
Training epochs                 & $1\mathrm{e}{2}$   & $5\mathrm{e}{1}$  \\
Batch size                      & $3.2\mathrm{e}{1}$ & $3.2\mathrm{e}{1}$  \\
Learning rate                   & $3\mathrm{e}{-5}$  & $3\mathrm{e}{-5}$   \\
$\beta_1$                       & $6\mathrm{e}{-1}$  & $2.5\mathrm{e}{-1}$ \\
$\beta_2$                       & $9\mathrm{e}{-1}$  & $9.9\mathrm{e}{-1}$ \\
$L_2$ regularization weight     & $3\mathrm{e}{-4}$  & $7\mathrm{e}{-4}$   \\
Label smoothing                 & $5\mathrm{e}{-2}$  & $2\mathrm{e}{-1}$   \\
$\gamma$                        & $1\mathrm{e}{-10}$ & $9\mathrm{e}{-9}$   \\
$\lambda$                       & $5\mathrm{e}{-2}$  & $3\mathrm{e}{-5}$   \\
Gradient clipping threshold     & $2\mathrm{e}{2}$   & $2\mathrm{e}{2}$    \\ \hline
\end{tabular}
\end{table}

\subsection*{Appendix B: Definition of model-based statistical dependency}

Given two random variables $X$ and $Y$ observed in pairs, we are interested in testing the hypothesis $P(Y|X) \neq P(Y)$, i.e. that $Y$ \emph{is}  dependent on $X$. We then use as evidence the performance one can achieve when trying to predict $Y$ from $X$ using a class of predictors $\mathcal{H}$ where $h \in \mathcal{H}: supp(X) \mapsto supp(Y)$. A test statistics is then given by:

\begin{equation}
\label{eq:test_statistics}
\min_{h \in \mathcal{H}} \mathbb{E}_{(x,y)\sim (X, Y)}\ell(h(x),y),    
\end{equation}
where $\ell:supp(Y)^2 \mapsto \mathbb{R}^+$ is a given loss function used to assess prediction performance.

We remark that the test result is of course dependent on the choice of $\mathcal{H}$ (hence the \emph{model-based} term) in the sense that a weak prediction performance might be simply due to a too constrained class of predictors taken into account rather than actual independence between variables of interest. However, reaching a strong performance using a simple enough $\mathcal{H}$ (e.g. hyperplanes) serves as strong evidence of statistical dependency. We further highlight that the $\mathcal{H}$-divergence \citep{ben2007analysis} is a particular case of the test statistics stated in \ref{eq:test_statistics} for the case where $Y$ indicates domain labels, and $\ell$ is the 0-1 loss.

\subsection*{Appendix C: Extra dependency analysis using FiLM parameters}

We extend the results reported in Tables \ref{tab:pacs_domainclassification} and \ref{tab:pacs_taskclassification} to understand how FiLM parameters are affected by properties of $z$ as well as how the conditioning procedure acts within the main model in order to induce the desired conditioning effect. We thus repeat linear classification experiments reported in section \ref{sec:z_properties}, and in this case we use FiLM parameters as inputs to the classifiers\footnote{We feed $z$ into FiLM layers and concatenated scaling and offset parameters are used to perform the experiments.}. Results are reported in Tables \ref{tab:pacs_domainclassification_film} and \ref{tab:pacs_taskclassification_film} for domain and task labels prediction, respectively. We observe from the results that the conclusions drawn in the case where $z$ is used as input to classifiers consistently transfer to the case where FiLM parameters are used, i.e. stronger domain/class dependency is observed for the cases of domain-consitional/self-modulated models. Given that, in the case of convolutional models, FiLM parameters determine which feature maps are to be given more or less importance, we claim a masking mechanism similar to cases such as mixture-of-experts \citep{shazeer2017outrageously,fedus2021switch} is what enforces domain-dependency in $M_{task}$, with differences lying in the facts that a soft masking happens in this case and feature maps are masked as opposed to weights.

\subsection*{Appendix D: Visualizing the conditioning variable $z$}

\begin{figure}[h]
\centering
\begin{subfigure}{.5\textwidth}
  \centering
  \includegraphics[width=.65\linewidth]{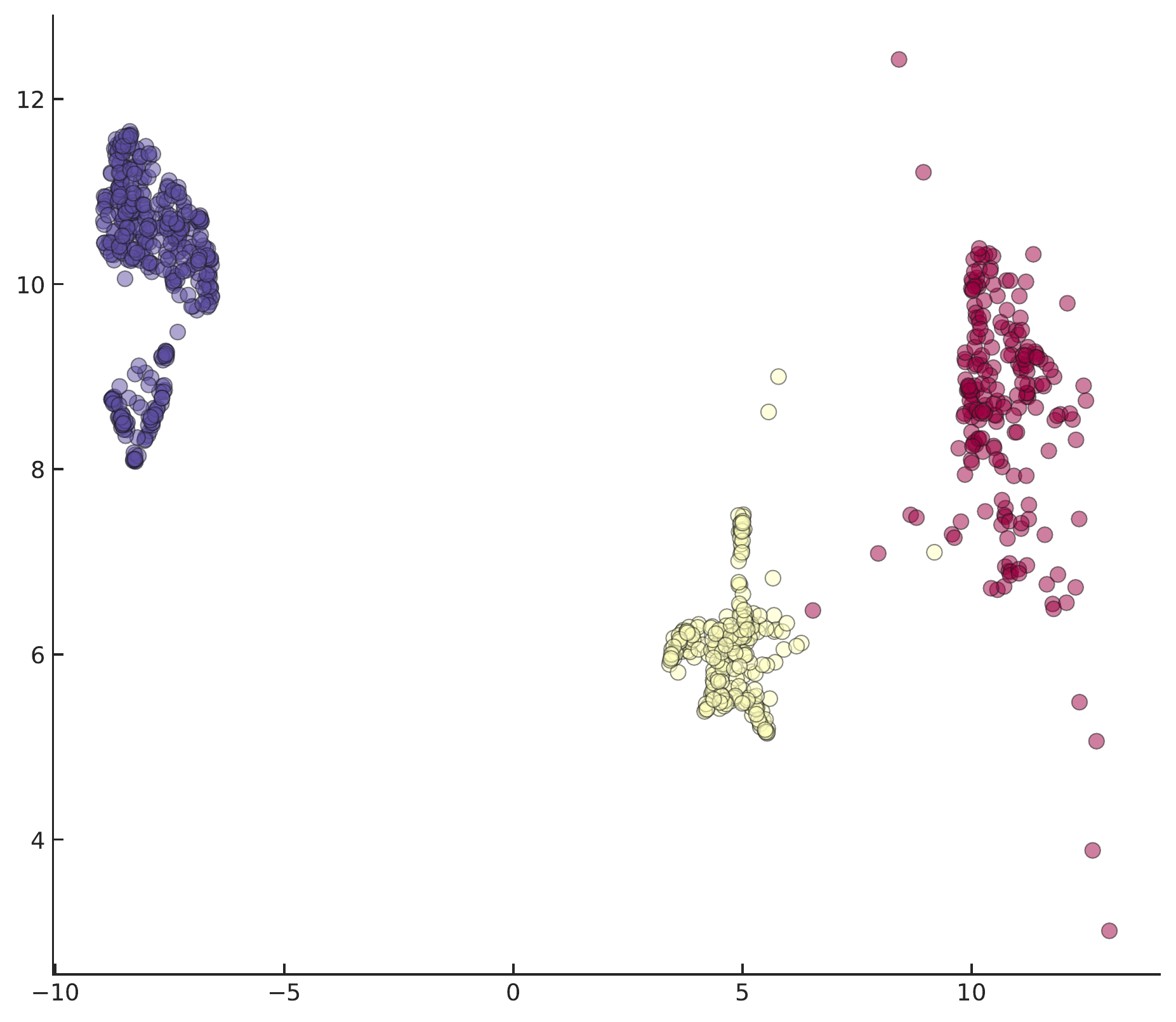}
  \caption{Test domain: Photo.}
  \label{fig:photo}
\end{subfigure}%
\begin{subfigure}{.5\textwidth}
  \centering
  \includegraphics[width=.65\linewidth]{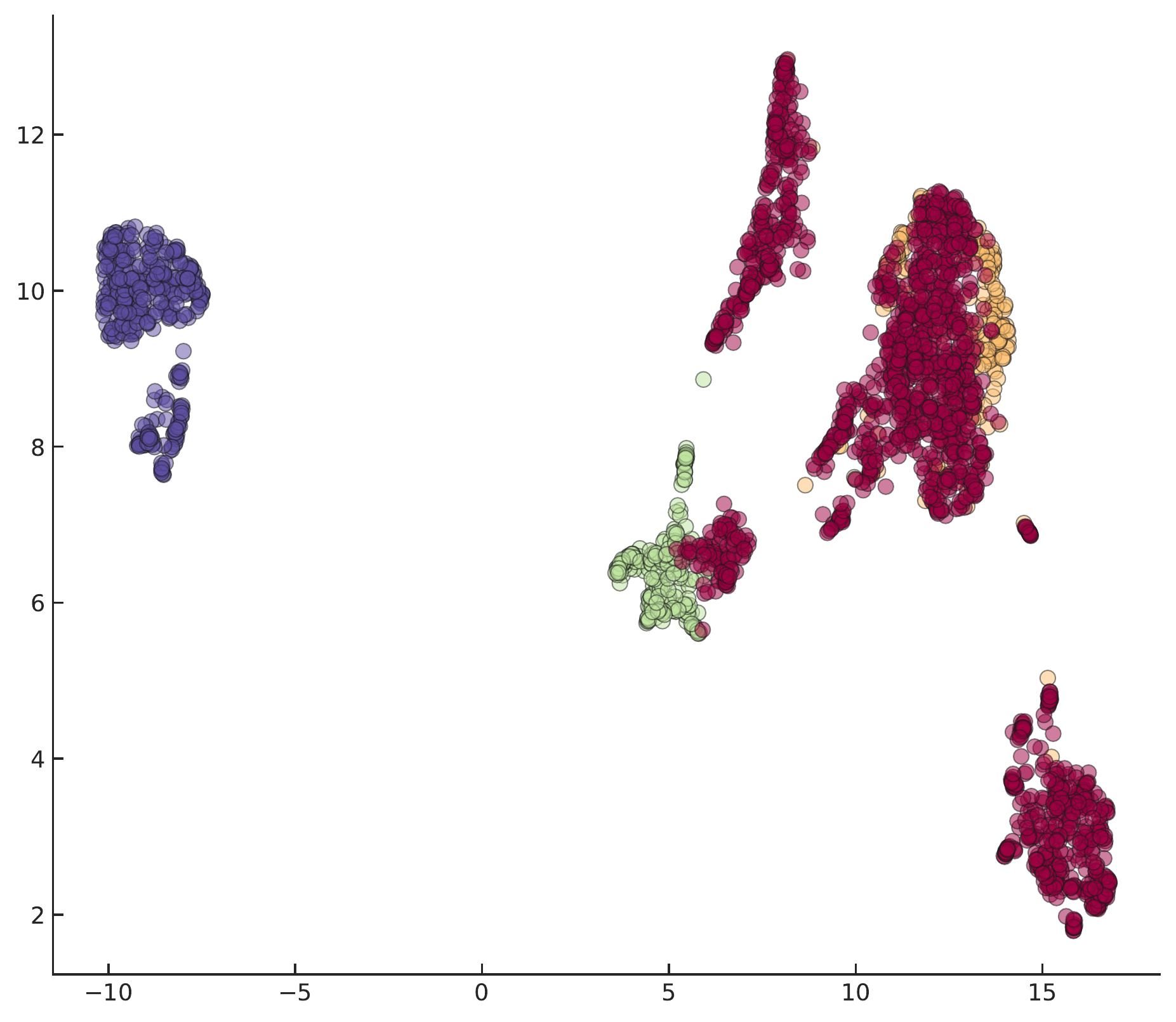}
  \caption{Test domain: Photo (included).}
  \label{fig:photo_test}
\end{subfigure}
\begin{subfigure}{.5\textwidth}
  \centering
  \includegraphics[width=.65\linewidth]{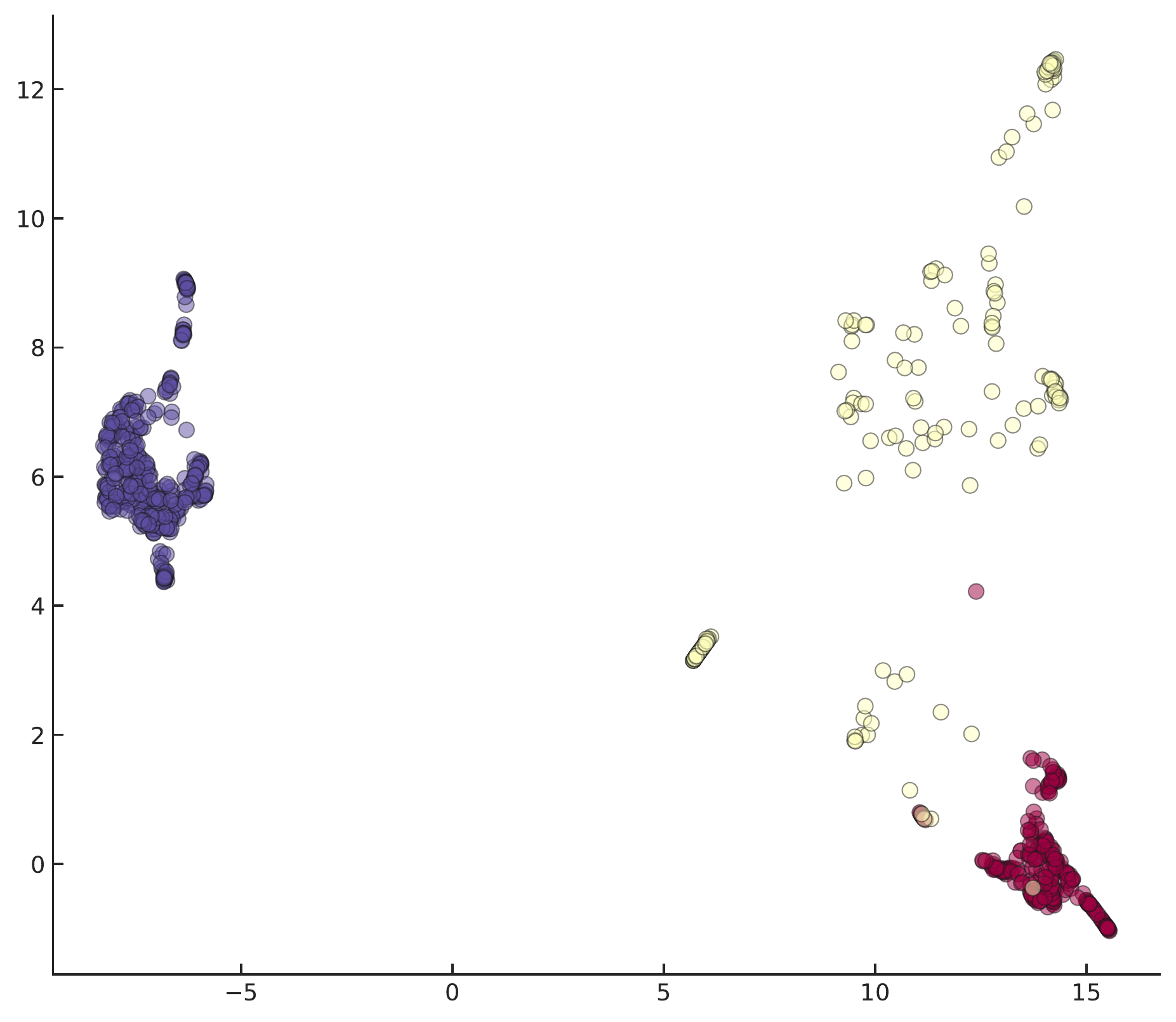}
  \caption{Test domain: Art painting.}
  \label{fig:art}
\end{subfigure}%
\begin{subfigure}{.5\textwidth}
  \centering
  \includegraphics[width=.65\linewidth]{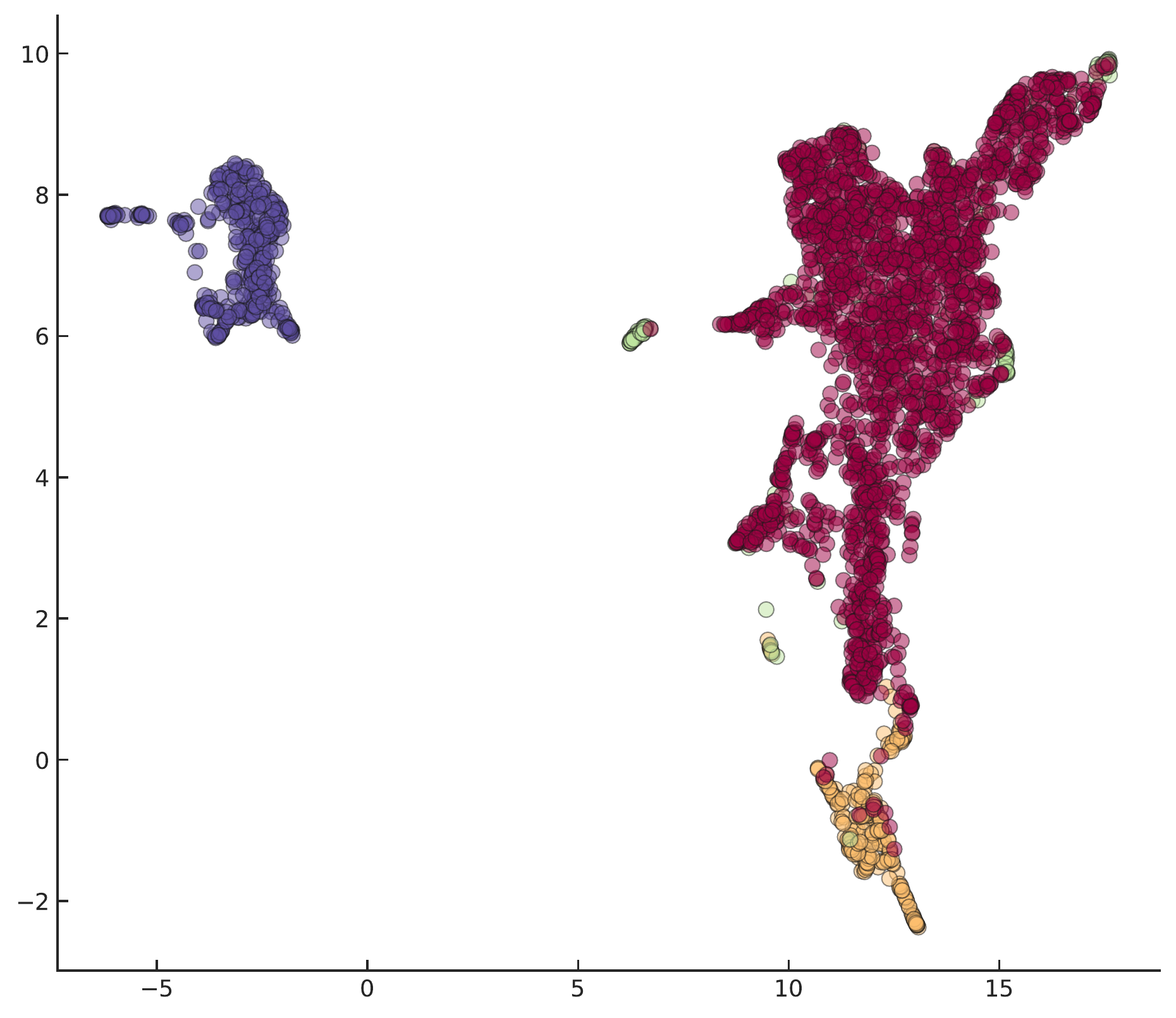}
  \caption{Test domain: Art paint. (included).}
  \label{fig:art_test}
\end{subfigure}
\begin{subfigure}{.5\textwidth}
  \centering
  \includegraphics[width=.65\linewidth]{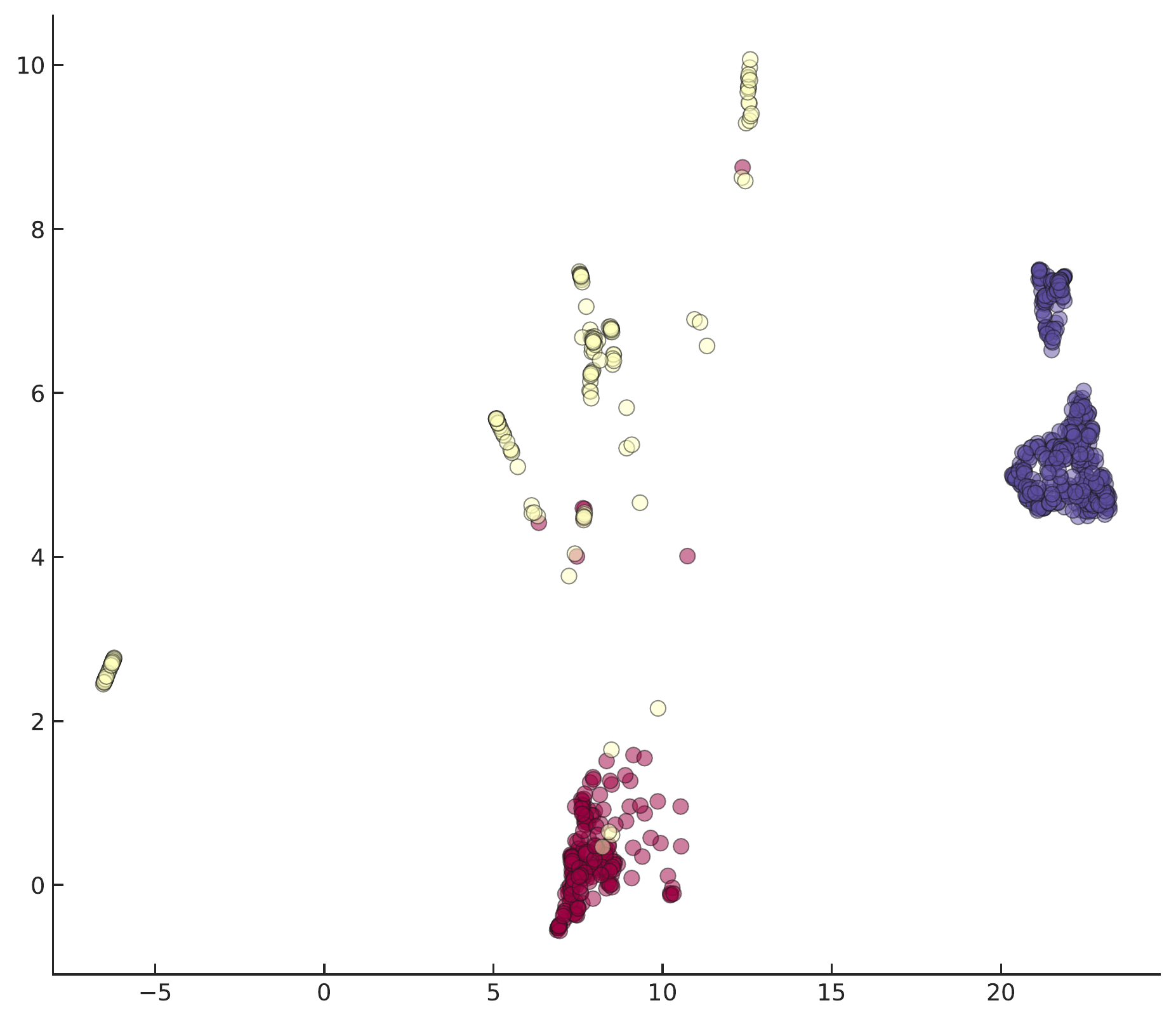}
  \caption{Test domain: Cartoon.}
  \label{fig:cartoon}
\end{subfigure}%
\begin{subfigure}{.5\textwidth}
  \centering
  \includegraphics[width=.65\linewidth]{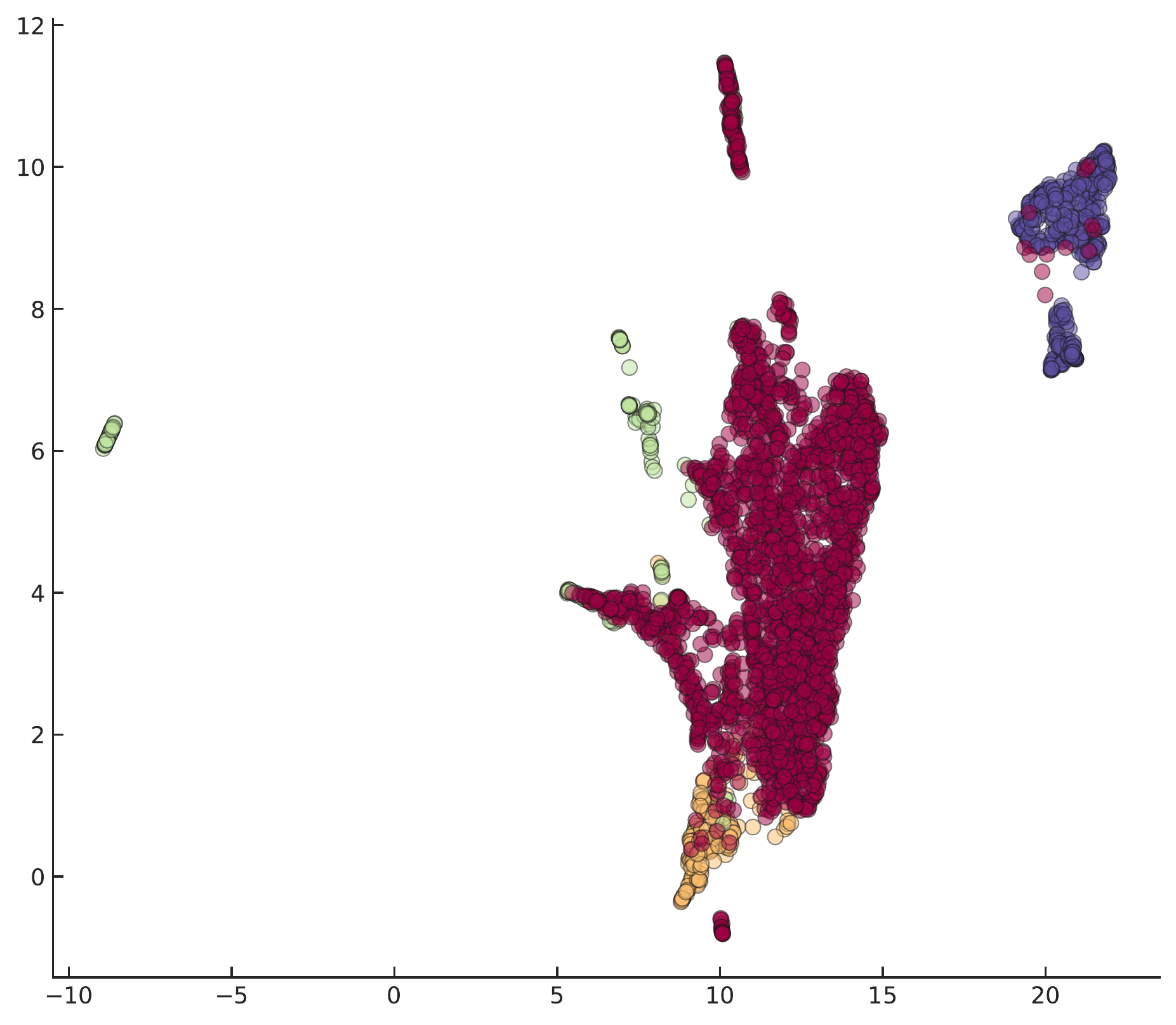}
  \caption{Test domain: Cartoon (included).}
  \label{fig:cartoon_test}
\end{subfigure}
\begin{subfigure}{.5\textwidth}
  \centering
  \includegraphics[width=.65\linewidth]{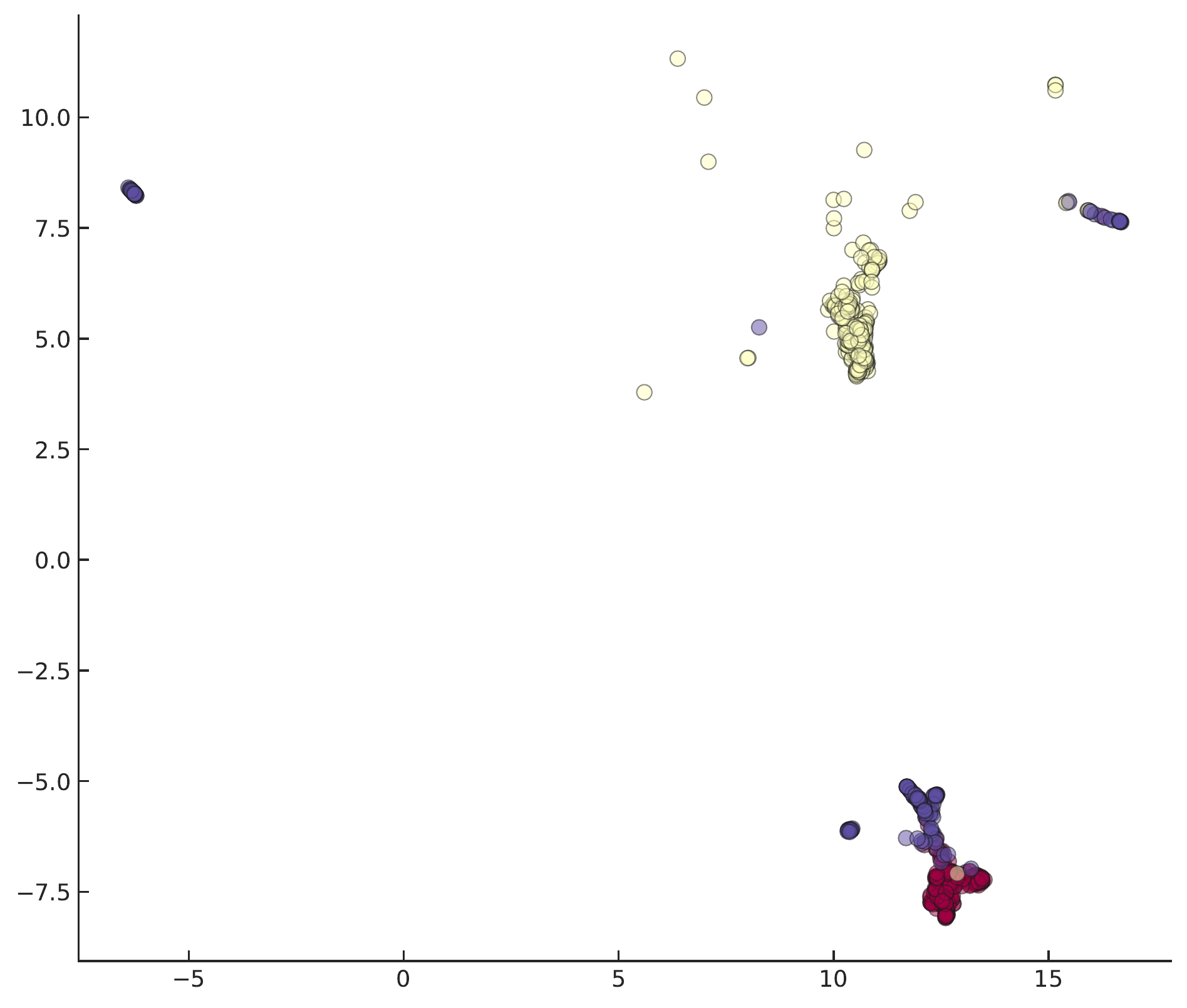}
  \caption{Test domain: Sketch.}
  \label{fig:sketch}
\end{subfigure}%
\begin{subfigure}{.5\textwidth}
  \centering
  \includegraphics[width=.65\linewidth]{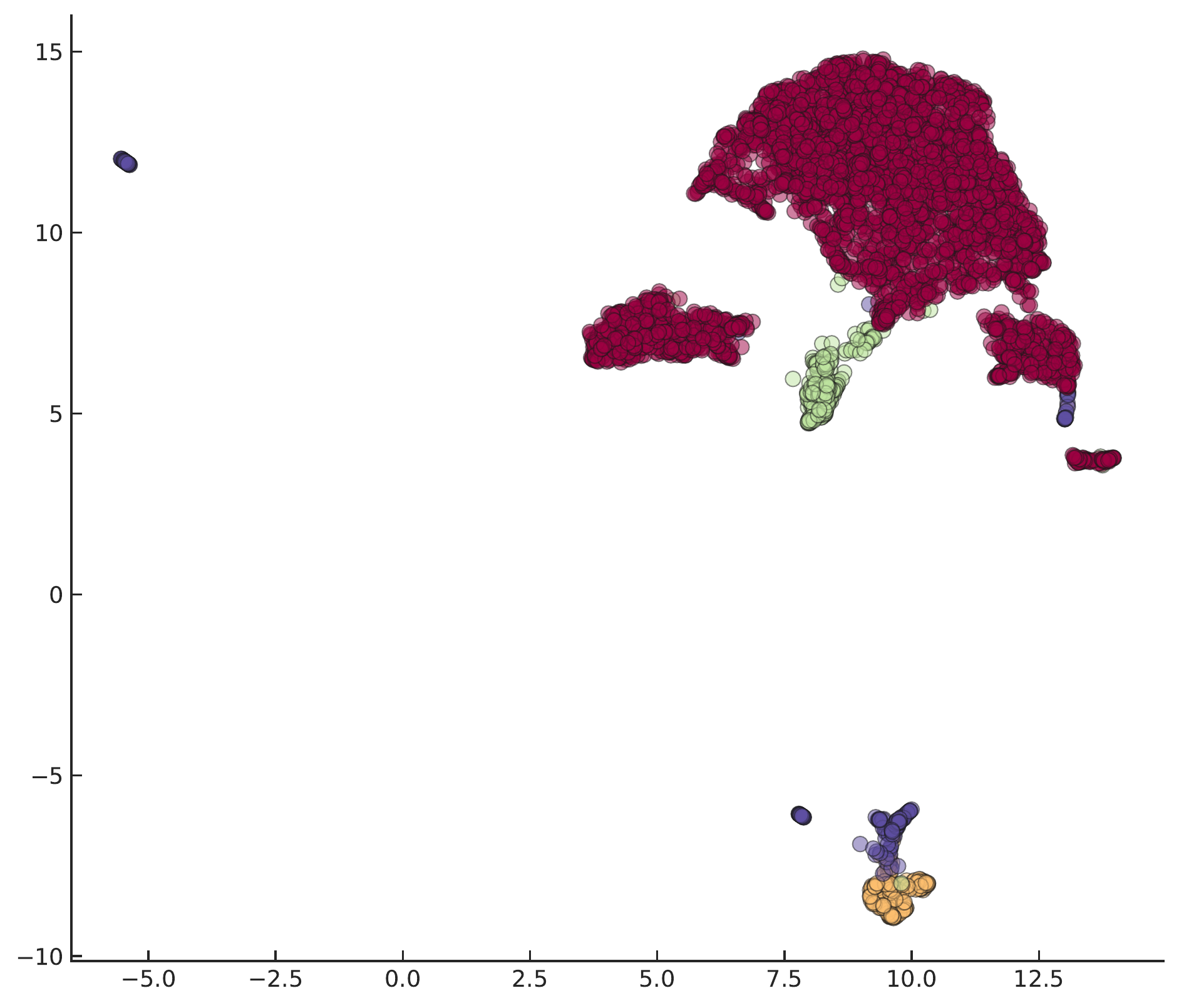}
  \caption{Test domain: Sketch (included).}
  \label{fig:sketch_test}
\end{subfigure}
\caption{UMAP projections of $z=M_{domain}(x)$ for conditional models trained on PACS.}
\end{figure}

\begin{figure}
\centering
\begin{subfigure}{.5\textwidth}
  \centering
  \includegraphics[width=.65\linewidth]{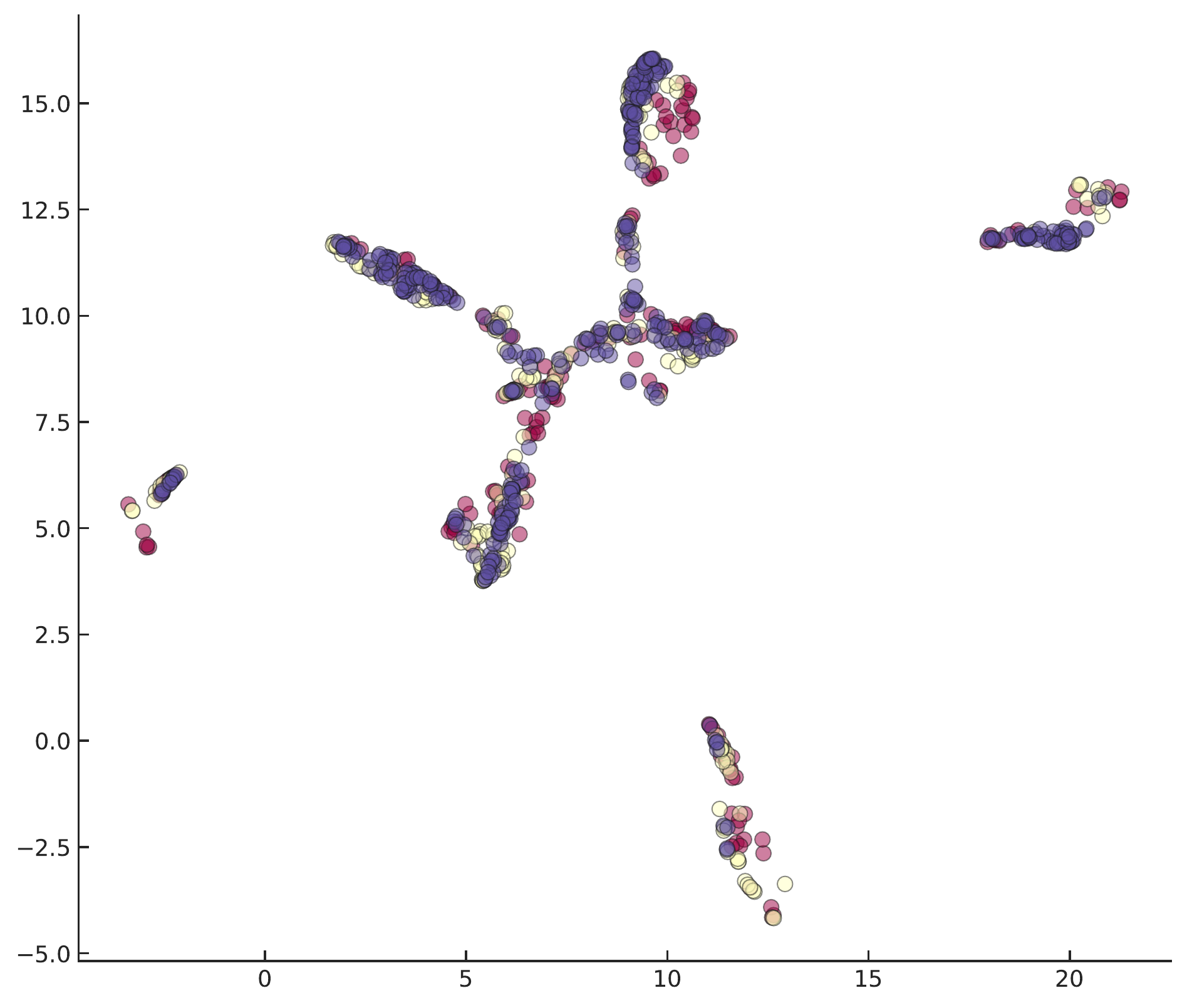}
  \caption{Test domain: Photo.}
  \label{fig:photo_self}
\end{subfigure}%
\begin{subfigure}{.5\textwidth}
  \centering
  \includegraphics[width=.65\linewidth]{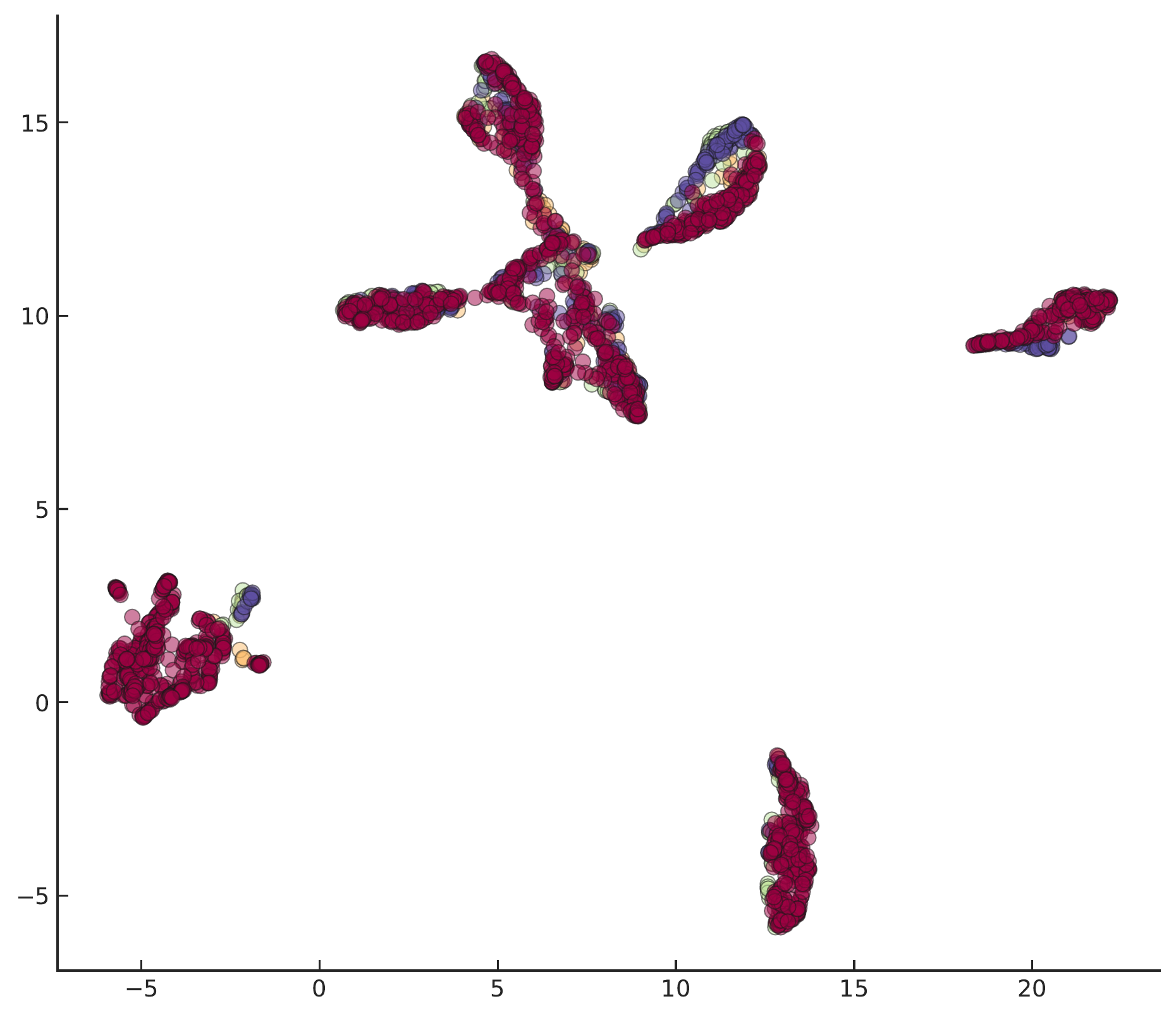}
  \caption{Test domain: Photo (included).}
  \label{fig:photo_test_self}
\end{subfigure}
\begin{subfigure}{.5\textwidth}
  \centering
  \includegraphics[width=.65\linewidth]{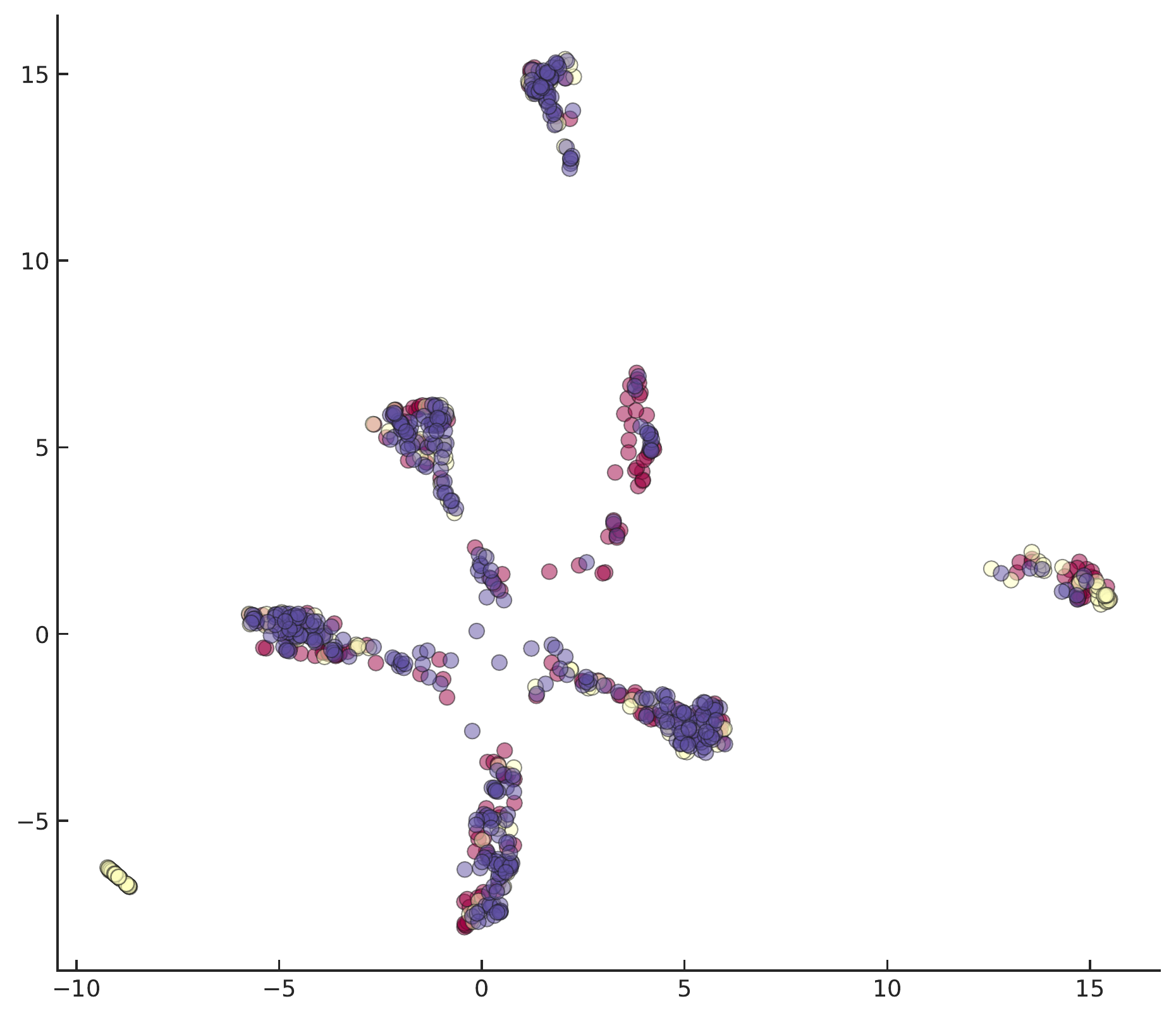}
  \caption{Test domain: Art painting.}
  \label{fig:art_self}
\end{subfigure}%
\begin{subfigure}{.5\textwidth}
  \centering
  \includegraphics[width=.65\linewidth]{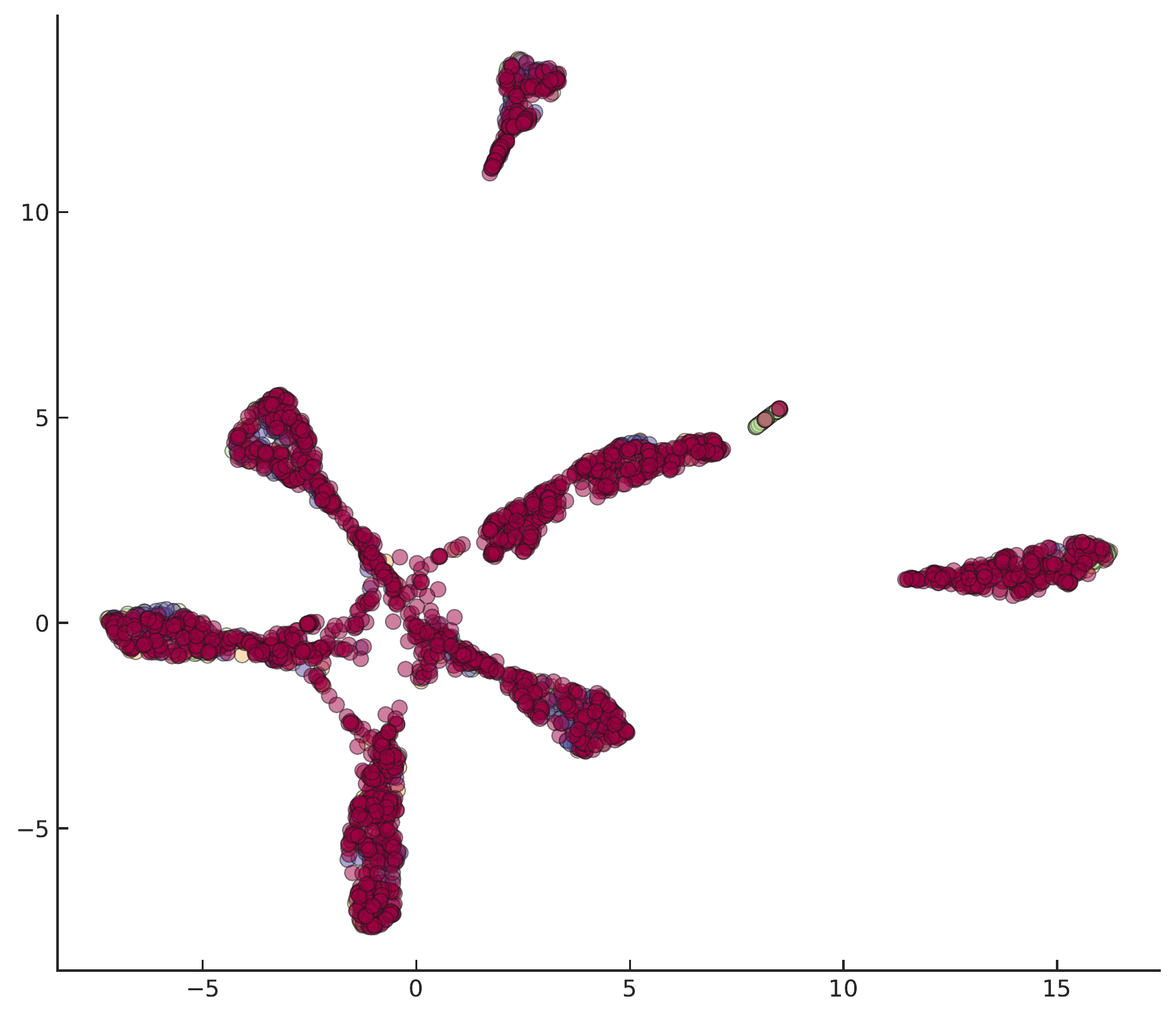}
  \caption{Test domain: Art paint. (included).}
  \label{fig:art_test_self}
\end{subfigure}
\begin{subfigure}{.5\textwidth}
  \centering
  \includegraphics[width=.65\linewidth]{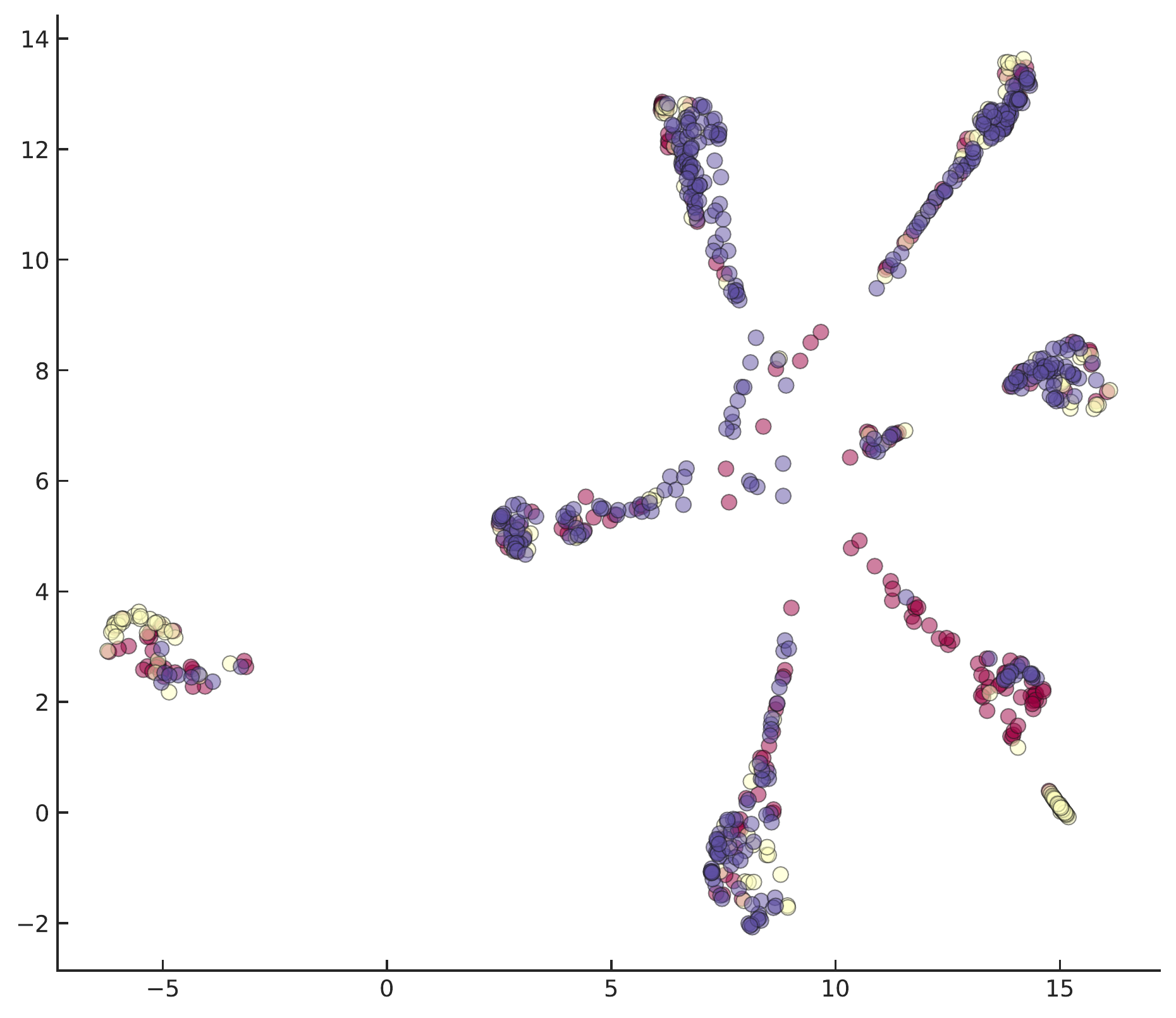}
  \caption{Test domain: Cartoon.}
  \label{fig:cartoon_self}
\end{subfigure}%
\begin{subfigure}{.5\textwidth}
  \centering
  \includegraphics[width=.65\linewidth]{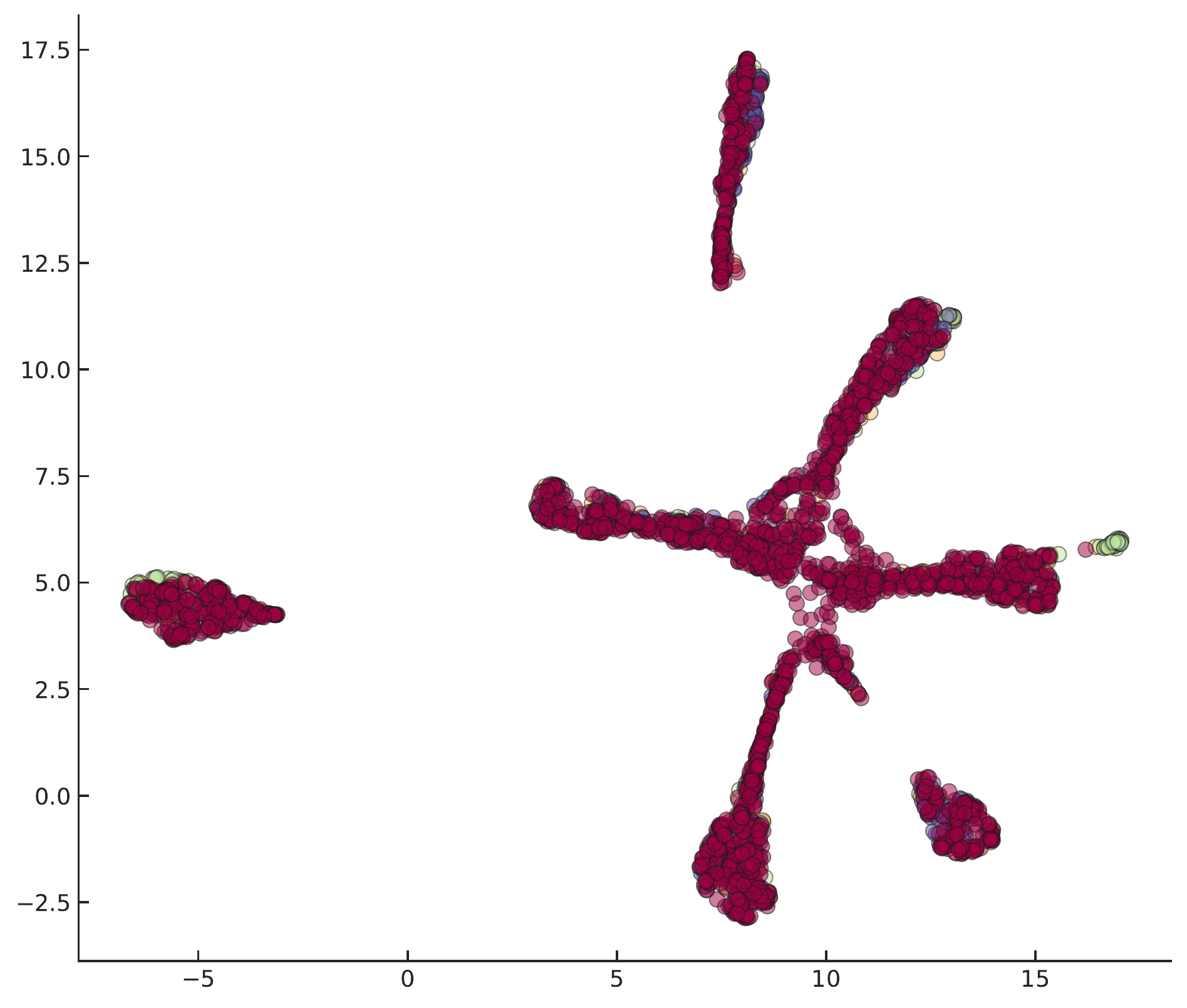}
  \caption{Test domain: Cartoon (included).}
  \label{fig:cartoon_test_self}
\end{subfigure}
\begin{subfigure}{.5\textwidth}
  \centering
  \includegraphics[width=.65\linewidth]{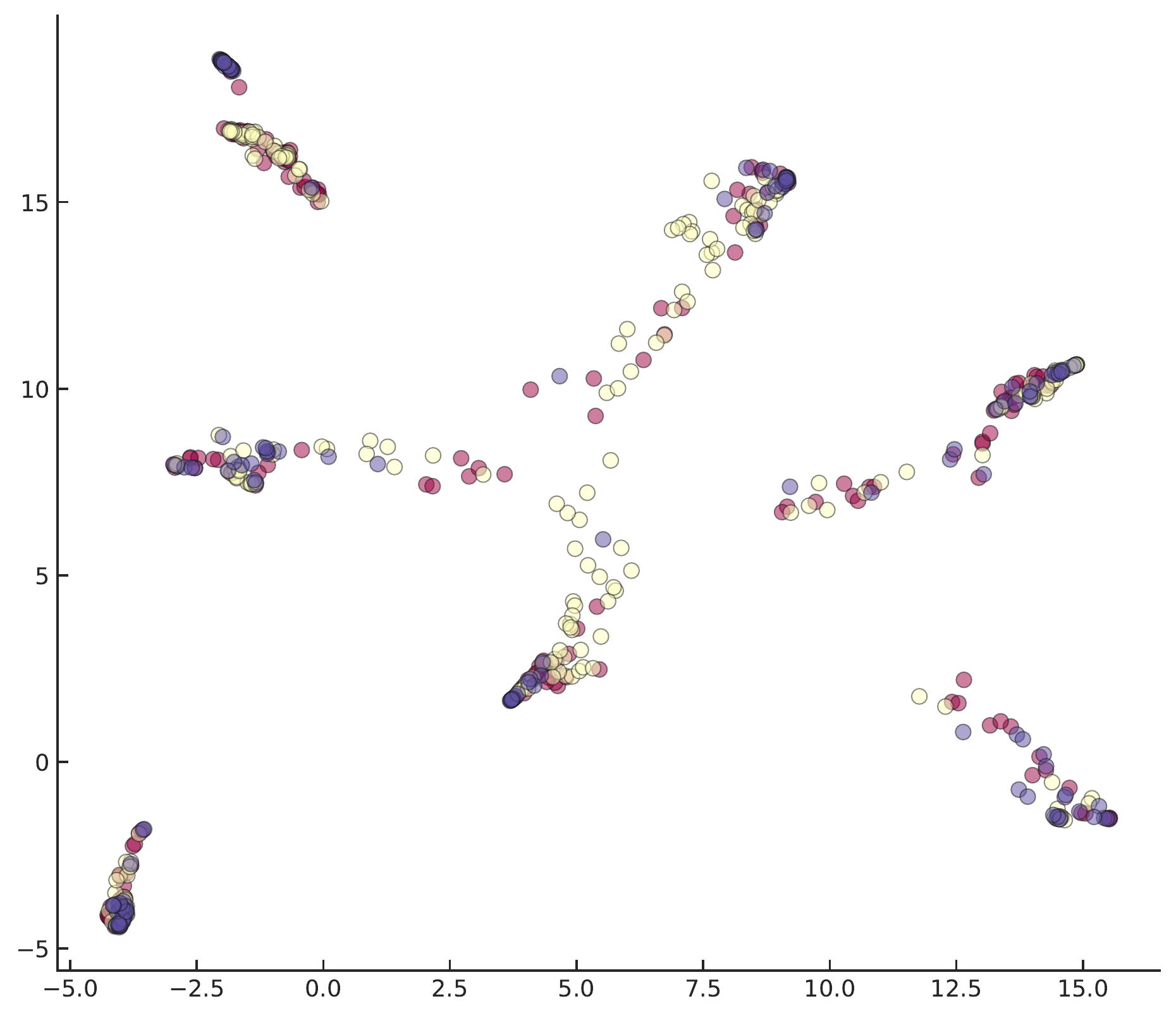}
  \caption{Test domain: Sketch.}
  \label{fig:sketch_self}
\end{subfigure}%
\begin{subfigure}{.5\textwidth}
  \centering
  \includegraphics[width=.65\linewidth]{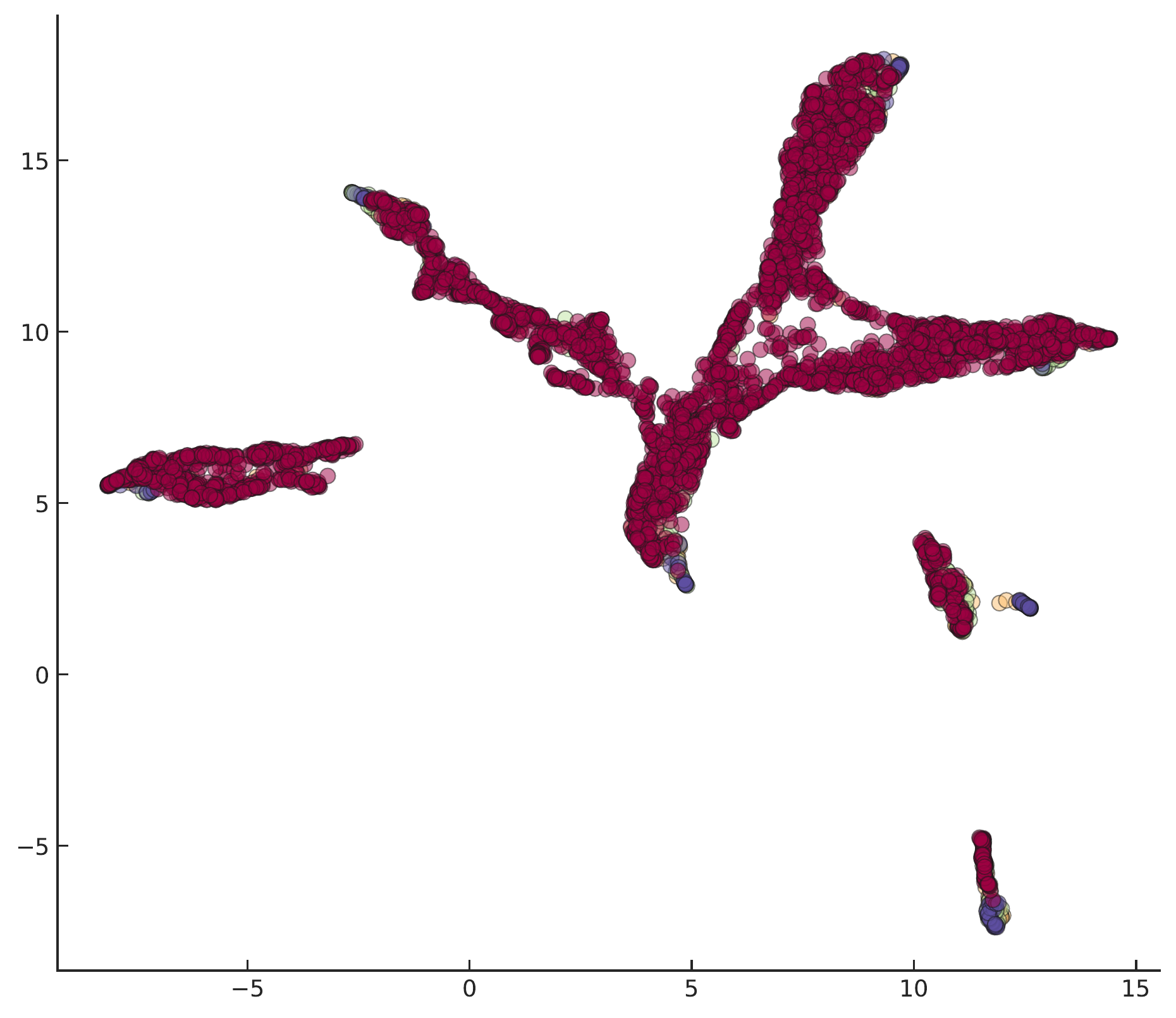}
  \caption{Test domain: Sketch (included).}
  \label{fig:sketch_test_self}
\end{subfigure}
\caption{UMAP projections of $z=M_{domain}(x)$ for self-modulated models trained on PACS.}
\end{figure}

Low dimensional projections of the conditioning variable $z=M_{domain}(x)$ obtained through UMAP \citep{mcinnes2018umap} are shown in Figures \ref{fig:photo}-\ref{fig:sketch_test} for the case of conditional models and in Figures \ref{fig:photo_self}-\ref{fig:sketch_test_self} for the case of self-modulated predictors. Pairs of images organized horizontally correspond to results obtained with the same models, i.e. when the same domain is left-out of training, and differ only in that that plots placed in the column on the right include representations from the test domain. Plots offer further evidence showing that domain supervision does induce domain-dependency in $z$. For the self-modulated case the domain-dependency is less evident from the projected $z$, but can still be verified once the full dimensional $z$ is used to train domain predictors, as reported in Table \ref{tab:pacs_domainclassification}.

\begin{table}[]
\centering
\caption{Prediction accuracy (\%) of logistic regression for performing domain classification on top of FiLM parameters. Results correspond to 95\% confidence intervals obtained through 5-fold cross validation.
\label{tab:pacs_domainclassification_film}}
\begin{tabular}{ccccc}
\hline
               & \multicolumn{4}{c}{\textbf{Left-out domain}}                                \\ \cline{2-5} 
               & \textbf{Art painting} & \textbf{Cartoon} & \textbf{Photo} & \textbf{Sketch} \\ \hline
$FiLM^1$       &                       &                  &                &                 \\ \hline
Conditional    & 99.01$\pm$0.19           & 98.97$\pm$0.20      & 99.64$\pm$0.11    & 96.10$\pm$0.55     \\
Self-modulated & 77.04$\pm$1.25           & 82.24$\pm$0.65      & 88.61$\pm$0.53    & 65.92$\pm$1.48     \\ \hline
$FiLM^2$       &                       &                  &                &                 \\ \hline
Conditional    & 98.63$\pm$0.18           & 98.33$\pm$0.20      & 99.64$\pm$0.11    & 95.78$\pm$0.61     \\
Self-modulated & 74.44$\pm$1.27           & 84.42$\pm$0.82      & 90.51$\pm$0.52    & 65.58$\pm$0.84     \\ \hline
$FiLM^3$       &                       &                  &                &                 \\ \hline
Conditional    & 98.76$\pm$0.27           & 98.20$\pm$0.10      & 99.64$\pm$0.19    & 96.26$\pm$0.56     \\
Self-modulated & 74.33$\pm$1.34           & 83.01$\pm$0.99      & 91.46$\pm$0.68    & 63.48$\pm$1.27     \\ \hline
$FiLM^4$       &                       &                  &                &                 \\ \hline
Conditional    & 98.26$\pm$0.36           & 98.58$\pm$0.29      & 99.41$\pm$0.21    & 95.62$\pm$0.52     \\
Self-modulated & 75.93$\pm$0.62           & 84.81$\pm$1.46      & 91.22$\pm$0.17    & 63.79$\pm$1.90     \\ \hline
\end{tabular}
\end{table}

\begin{table}[]
\centering
\caption{Prediction accuracy (\%) of logistic regression for predicting class labels using FiLM parameters. Results correspond to 95\% confidence intervals obtained through 5-fold cross validation.
\label{tab:pacs_taskclassification_film}}
\begin{tabular}{ccccc}
\hline
               & \multicolumn{4}{c}{\textbf{Left-out domain}}                                \\ \cline{2-5} 
               & \textbf{Art painting} & \textbf{Cartoon} & \textbf{Photo} & \textbf{Sketch} \\ \hline
$FiLM^1$       &                       &                  &                &                 \\ \hline
Conditional    & 78.79$\pm$0.35           & 77.47$\pm$2.11      & 68.56$\pm$0.75    & 74.84$\pm$0.46     \\
Self-modulated & 94.30$\pm$0.67           & 92.54$\pm$1.16      & 91.34$\pm$1.08    & 93.51$\pm$0.57     \\ \hline
$FiLM^2$       &                       &                  &                &                 \\ \hline
Conditional    & 79.53$\pm$0.87           & 78.12$\pm$0.68      & 66.54$\pm$1.38    & 75.16$\pm$0.59     \\
Self-modulated & 94.54$\pm$0.32           & 92.15$\pm$0.37      & 90.98$\pm$0.54    & 92.21$\pm$1.14     \\ \hline
$FiLM^3$       &                       &                  &                &                 \\ \hline
Conditional    & 78.29$\pm$0.83           & 76.57$\pm$1.55      & 70.23$\pm$0.72    & 75.00$\pm$0.16     \\
Self-modulated & 94.54$\pm$0.65           & 92.92$\pm$0.53      & 91.22$\pm$0.70    & 93.02$\pm$0.66     \\ \hline
$FiLM^4$       &                       &                  &                &                 \\ \hline
Conditional    & 79.53$\pm$0.51           & 77.35$\pm$0.72      & 69.04$\pm$1.51    & 75.17$\pm$1.30     \\
Self-modulated & 94.29$\pm$0.56           & 92.41$\pm$0.71      & 90.50$\pm$1.03    & 92.37$\pm$0.59     \\ \hline
\end{tabular}
\end{table}

\end{document}